\documentclass[sigconf]{acmart}
\AtBeginDocument{%
  }

\setcopyright{none}           
\renewcommand\footnotetextcopyrightpermission[1]{} 
\usepackage{times}
\usepackage{latexsym}
\usepackage{amsmath}
\usepackage{subcaption}
\usepackage{colortbl}
\usepackage{pifont}
\usepackage{diagbox}
\usepackage{tablefootnote}
\usepackage{threeparttable}

\usepackage{hyperref}
\usepackage{url}
\usepackage{graphicx}
\usepackage{verbatim}
\usepackage{fancyvrb}
\usepackage{listings}
\usepackage{booktabs}
\usepackage{array}  
\usepackage{xcolor}   
\usepackage{cleveref}

\usepackage{longtable}
\usepackage{multirow}
\usepackage{tabularx}
\usepackage{ragged2e}
\usepackage{makecell}
\usepackage{enumitem}
\usepackage[smartEllipses]{markdown}
\usepackage{booktabs}
\usepackage{longtable}
\usepackage{tcolorbox}
\usepackage{enumitem}

\usepackage{amssymb}
\usepackage{multicol}

\usepackage[linesnumbered,ruled,vlined]{algorithm2e}

\newcommand{\True}{\textbf{True}}
\newcommand{\False}{\textbf{False}}

\SetKwProg{Fn}{def}{:}{} 
\SetKw{KwIn}{in}         
\SetKw{KwNot}{not}       
\DontPrintSemicolon      

\definecolor{graybg}{gray}{0.95}

\begin{document}

\title{FinRule-Bench: A Benchmark for Joint Reasoning over Financial Tables and Principles}

\author{Arun Vignesh Malarkkan}
\email{arun.malarkkan@asu.edu}
\affiliation{%
  \institution{Arizona State University}
  \city{Tempe}
  \state{Arizona}
  \country{USA}
}

\author{Manan Roy Choudhury}
\email{mroycho1@asu.edu}
\affiliation{%
  \institution{Arizona State University}
  \city{Tempe}
  \state{Arizona}
  \country{USA}
}

\author{Guangwei Zhang}
\email{changhu1633@gmail.com}
\affiliation{%
 \institution{Pine AI}
 \country{Singapore, Singapore}}

\author{Vivek Gupta}
\email{vgupt140@asu.edu}
\affiliation{%
  \institution{Arizona State University}
  \city{Tempe}
  \state{Arizona}
  \country{USA}}

\author{Qingyun Wang}
\email{qwang16@wm.edu}
\affiliation{%
  \institution{The College of William \& Mary}
  \city{Williamsburg}
  \state{Virginia}
  \country{USA}}

\author{Yanjie Fu}
\email{yanjie.fu@asu.edu}
\affiliation{%
  \institution{Arizona State University}
  \city{Tempe}
  \state{Arizona}
  \country{USA}
}

\author{Denghui Zhang}
\email{dzhang42@stevens.edu}
\affiliation{%
  \institution{Stevens Institute of Technology}
  \state{New Jersey}
  \country{USA}
}

\renewcommand{\shortauthors}{Malarkkan, A.V., et al.}

\begin{abstract}
Large language models (LLMs) are increasingly applied to financial analysis, yet their ability to audit structured financial statements under explicit accounting principles remains poorly explored. Existing benchmarks primarily evaluate question answering, numerical reasoning, or anomaly detection on synthetically corrupted data, making it unclear whether models can reliably verify or localize rule compliance on correct financial statements.
We introduce \textbf{FinRule-Bench}, a benchmark for evaluating diagnostic completeness in rule-based financial reasoning over real-world financial tables. FinRule-Bench pairs ground-truth financial statements with explicit, human-curated accounting principles and spans four canonical statement types: Balance Sheets, Cash Flow Statements, Income Statements, and Statements of Equity. The benchmark defines three auditing tasks that require progressively stronger reasoning capabilities: (i) \textbf{rule verification}, which tests compliance with a single principle; (ii) \textbf{rule identification}, which requires selecting the violated principle from a provided rule set; and (iii) \textbf{joint rule diagnosis}, which requires detecting and localizing multiple simultaneous violations at the record level.
We evaluate LLMs under zero-shot and few-shot prompting, and introduce a causal-counterfactual reasoning protocol that enforces consistency between decisions, explanations, and counterfactual judgments. 
Across tasks and statement types, we find that while models perform well on isolated rule verification, performance degrades sharply for rule discrimination and multi-violation diagnosis. Errors are dominated by incomplete coverage and mislocalization of violations, revealing systematic failures that are not captured by existing financial or tabular benchmarks. FinRule-Bench provides a principled and reproducible testbed for studying rule-governed reasoning, diagnostic coverage, and failure modes of LLMs in high-stakes financial analysis. We release the FinRule-Bench benchmark along with the complete evaluation codebase, including task-specific LLM prompting scripts for all three tasks across all financial statement types, at \url{https://anonymous.4open.science/r/FinRule-Bench-6723/}.

\end{abstract}


\maketitle

\vspace{1cm}
\section{Introduction}

Large language models (LLMs) are increasingly deployed for financial analysis and decision support, including applications such as numerical question answering, report summarization, and anomaly detection. However, a central task in financial auditing remains poorly understood from the perspective of LLM evaluation: determining whether structured financial statements comply with explicit accounting principles and, if not, identifying and localizing the precise sources of non-compliance. This requirement goes beyond producing correct answers or detecting isolated inconsistencies; it demands \emph{diagnostic completeness}, i.e., the ability to exhaustively enforce a formal rule system over structured, interdependent financial tables.

Financial reporting underpins modern capital markets and serves as the primary mechanism through which firms communicate performance, risk, and regulatory compliance to investors and oversight bodies. The reliability of these reports is enforced through rigorous auditing processes overseen by regulatory authorities such as the PCAOB~\citep{pcaob:website} and grounded in formal accounting frameworks including U.S. GAAP~\citep{fasb:website} and IFRS~\citep{iasb:website}. Auditing requires verifying that financial statements satisfy a large set of explicit and interdependent accounting principles, including arithmetic identities, hierarchical aggregation constraints, and conditional applicability rules. Effective auditing therefore requires structured reasoning that can be evaluated for completeness, requiring models to identify the violated principles and to attribute each violation localized to the specific records, rather than producing binary judgments or isolated answers.

\begin{figure*}[ht]
    \centering
    \includegraphics[width=0.9\linewidth]{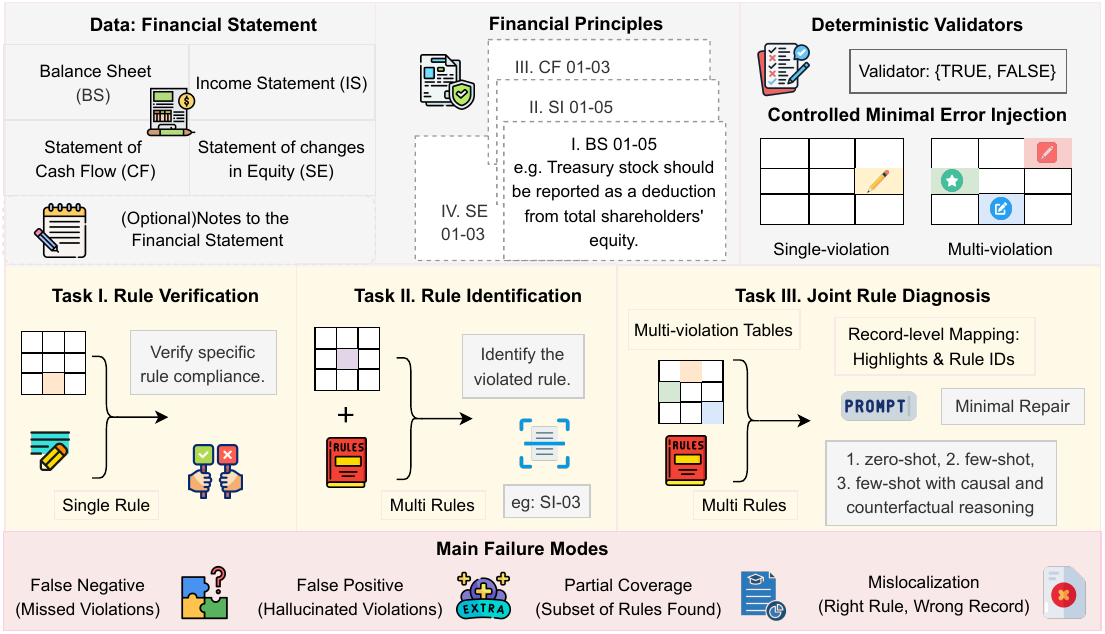}
    \caption{FinRule-Bench Benchmark Overview. FinRule-Bench evaluates LLMs on financial auditing tasks that require both intra-table reasoning and joint reasoning across multiple financial statements under explicit accounting rules.}
    \label{fig:overview}
    \vspace{-0.2cm}
\end{figure*}

Existing benchmarks only partially capture these requirements. General table question answering datasets, such as TAT-QA~\citep{zhu-etal-2021-tat}, primarily evaluate single-table extraction and numerical reasoning over individual tables, without requiring validation against external rule systems. Financial question answering benchmark,s including FinQA~\citep{chen2021finqa} and ConvFinQA~\citep{chen-etal-2022-convfinqa}, emphasize retrieval and arithmetic computation but do not assess whether models can enforce accounting principles or discriminate among competing rule violations. As a result, models can achieve high accuracy on existing benchmarks while remaining unsuitable for auditing, where correctness depends on exhaustive rule enforcement and precise attribution rather than isolated answers. 

This gap motivates a distinct evaluation setting. Verifying accounting compliance over ground-truth financial statements differs fundamentally from question answering or anomaly detection. In realistic auditing scenarios, financial tables are often correct by construction, and failures arise from incorrect reasoning rather than from corrupted or noisy data. Consequently, evaluating LLMs in this setting requires benchmarks that isolate reasoning fidelity from data artifacts and directly test whether models can exhaustively identify violated rules and attribute each violation to the appropriate records in structured financial reports.

To address this need, we introduce \textbf{FinRule-Bench}, a benchmark designed to evaluate rule-based financial reasoning over real-world financial statements. FinRule-Bench pairs structured financial tables with explicit, human-curated accounting principles and defines three auditing tasks that require progressively stronger reasoning capabilities: rule verification, which tests compliance with a single principle; rule identification, which requires selecting the uniquely violated principle from a provided rule set; and joint rule diagnosis, which requires detecting and localizing multiple simultaneous violations at the record level. Together, these tasks evaluate a model’s ability to compose accounting rules, distinguish violated principles, and localize violations to specific records under realistic auditing conditions.

Beyond task formulation, FinRule-Bench introduces a \emph{causal-counterfactual reasoning} protocol for structured financial analysis. This protocol requires models not only to output decisions, but also to justify them through explicit causal explanations and to assess whether minimal counterfactual modifications to the financial table would alter those decisions. Rather than aiming to improve accuracy uniformly, this protocol serves as a diagnostic instrument: by comparing decisions and counterfactual judgments, it reveals inconsistencies, incomplete coverage, and mislocalization errors that are not observable when evaluating final answers alone.

Overall, FinRule-Bench provides a principled testbed for evaluating rule-governed reasoning, diagnostic coverage, and failure modes of LLMs in high-stakes financial analysis. By focusing on rule compliance over correct financial statements, it complements existing tabular and financial benchmarks and enables precise analysis of where current models succeed and where they fundamentally fall short. We summarize our contributions as follows:
\vspace{-0.1cm}
\begin{enumerate}
    \item \textbf{Benchmark.} We introduce \textbf{FinRule-Bench}, a benchmark derived from real-world financial statements spanning Balance Sheets, Cash Flow Statements, Income Statements, and Statements of Equity. The benchmark pairs ground-truth tables with explicit accounting principles, enabling controlled evaluation of rule-based financial reasoning beyond question answering or anomaly detection.
    \item \textbf{Task Suite.} We formalize three complementary auditing tasks that evaluate progressively stronger reasoning capabilities: rule verification, which tests compliance with a single accounting principle; rule identification, which requires selecting the violated principle from a provided rule set; and joint rule diagnosis, which requires multi-violation detection and record-level localization.
    \item \textbf{Reasoning and Evaluation Framework.} We introduce a causal-counterfactual prompting protocol for structured financial reasoning and introduce diagnostic evaluation analyses, including rule-complexity breakdowns, error-type categorization, and performance-cost scalability studies, that reveal systematic failures in diagnostic completeness not captured by aggregate accuracy alone.
\end{enumerate}

\section{Related Work}

\noindent\textbf{LLMs for Table Understanding.}
Applying LLMs over tabular data poses challenges that differ fundamentally from textual reasoning, including modeling two-dimensional structure, performing precise numerical reasoning, and preserving semantic relationships across rows and columns~\citep{fang2024largelanguagemodelsllmstabular}. Early work addressed these challenges through task-specific, encoder-based models such as TaPas~\citep{herzig-etal-2020-tapas}, TaBERT~\citep{yin-etal-2020-tabert}, TURL~\citep{10.1145/3542700.3542709}, and TUTA~\citep{DBLP:conf/kdd/WangDJLFHZ21}, which explicitly modeled table structure and cell-level interactions. More recent approaches leverage instruction-tuned LLMs and prompting-based methods. Table-centric architectures like Table-GPT~\citep{10.1145/3654979} and TableLlama~\citep{zhang-etal-2024-tablellama} adapt LLMs for tabular reasoning, while agent-based frameworks including StructGPT~\citep{jiang-etal-2023-structgpt} and DATER~\citep{10.1145/3539618.3591708} incorporate external tools or structured representations. Across these approaches, serialization strategies and prompting techniques, including Chain-of-Thought reasoning~\citep{10.5555/3600270.3602070}, have been shown to substantially improve performance. Comprehensive surveys summarize tasks, datasets, and methodological trends in this area~\citep{fang2024largelanguagemodelsllmstabular,Lu_2025}. Despite these works, evaluation in tabular reasoning research remains dominated by question answering and cell-level prediction. These settings do not assess whether models can exhaustively identify and localize rule violations in complex, real-world tables, which is a central requirement in auditing and compliance-oriented applications.

\noindent\textbf{LLMs for Financial Data Processing and Analysis.}
LLMs have been widely applied to financial data processing tasks~\citep{Zhao2024RevolutionizingFW,lee2025large}, including algorithmic trading~\citep{10.1145/3677052.3698696}, investment analysis~\citep{fatouros2025marketsenseai20enhancingstock}, market forecasting, sentiment analysis~\citep{10638546}, fraud detection~\citep{10760895}, and risk assessment~\citep{10803746}. A parallel line of work focuses on domain-adapted financial language models, such as FinGPT~\citep{liu2023fingptdemocratizinginternetscaledata}, trained on large-scale financial corpora, which have been adapted to improve performance. Recent work has also explored Retrieval-Augmented Generation (RAG) to improve factual grounding and contextual awareness in financial analysis~\citep{wang2025financialanalysisintelligentfinancial}. Benchmark efforts such as FinDABench~\citep{liu-etal-2025-findabench} and AuditBench~\cite{wang2025auditbench} provide standardized evaluation settings for financial tasks involving retrieval, summarization, and prediction. However, these benchmarks largely evaluate information access or descriptive accuracy. They do not test whether models can apply explicit accounting principles to verify the internal consistency or diagnose violations in financial statements. Consequently, high performance on existing financial benchmarks does not imply suitability for auditing or compliance verification.

\noindent\textbf{Rule-Based and Logical Reasoning on Structured Data.}
Rule-based and logical reasoning over structured data has a long history in artificial intelligence~\citep{10.1145/3447772}, with recent work shifting toward neurosymbolic approaches~\citep{10.1007/s10462-023-10448-w}. Several methods augment LLMs with symbolic reasoning modules, constraint solvers, or logic-guided decoding to improve logical correctness and reasoning faithfulness~\citep{creswell2022selectioninferenceexploitinglargelanguage,lyu2023faithfulchainofthoughtreasoning}. Related efforts study complex query answering~\citep{xiao-cao-2024-complex}, multi-hop inference~\citep{kim-etal-2024-improving-multi}, and logic-guided neural architectures~\citep{10.1145/3748239.3748249}. Benchmark datasets such as LogicGame~\citep{gui2024logicgamebenchmarkingrulebasedreasoning,liu2023evaluatinglogicalreasoningability} provide insight into abstract reasoning ability but rely on synthetic rules with limited domain semantics. In contrast, FinRule-Bench evaluates rule-based reasoning grounded in real financial statements and formal accounting principles, where violations correspond to realistic auditing errors and introduce challenges such as rule prioritization, attribution, and multi-violation coverage not captured by existing benchmarks.

\section{FinRule-Bench: Dataset and Evaluation Framework}

Motivated by the limitations of existing tabular financial benchmarks, we introduce \textbf{FinRule-Bench}, a benchmark for evaluating rule-based reasoning over financial tables under explicit accounting principles. FinRule-Bench reflects the core demands of financial auditing, where correctness depends on verifying compliance between multiple interdependent rules and structured financial statements. Unlike prior benchmarks that emphasize question answering or anomaly detection on synthetic corrupted data, FinRule-Bench evaluates reasoning over ground-truth financial data and focuses on rule-based reasoning fidelity as the primary objective. 

FinRule-Bench is built with real-world financial statements extracted from corporate filings. This ensures that the evaluation is focused on realistic tabular data and not on synthesized datasets. Correctness of data is validated by explicit, human-curated accounting principles and is implemented as deterministic validators, making the ground truth reliable and reproducible. To evaluate the LLMs on this benchmark, we curate a controlled, rule-aware error injection to those clean financial statements at record-level. With the clean tabular financial statements as a ground-truth base dataset, the error-injected dataset instances are created via controlled, minimal edits that violate specific accounting principles. The task definitions correspond to different constrained perspectives over this perturbed dataset, and differ only in violation cardinality, required model outputs, and complexity.
In this section, we describe dataset construction with statement-specific rules and validators, task formulation, evaluation protocols with defined prompting techniques, and dataset statistics.

\vspace{-0.2cm}
\subsection{Dataset Construction: Base Financial Statements}

FinRule-Bench is constructed from real-world corporate financial disclosures, with 2024 Form 10-K filings as the primary data source (\Cref{fig:financial_statement}). These are standardized, regulator-approved financial filings that adhere to formal accounting frameworks and principles.
Financial statements are extracted from filings and converted into structured \emph{Markdown tables}. We use the Qwen-2.5-VL-72B vision-language model~\citep{bai2025qwen25vltechnicalreport} to assist in transcribing visual financial statements into Markdown tabular structures while preserving row hierarchy, column semantics, and numerical values (\Cref{table:table_samples}). Markdown serves as the canonical representation because it is both human-readable and directly consumable by large language models.
Each record in the markdown file corresponds to a single company financial statement for a given fiscal year. The statement types include Balance Sheets, Cash Flow Statements, Income Statements, and Statements of Equity. When a statement of a specific company in a single fiscal year spans multiple pages or images, all fragments are merged into a single coherent Markdown record. Each finalized table is stored as an individual file indexed by statement type.
All tables included in FinRule-Bench correspond to ground-truth financial statements. No numerical values are modified, perturbed, or synthetically corrupted during dataset construction. As a result, any detected violations arise exclusively from evaluating the tables against explicit accounting principles rather than from data noise or extraction artifacts.

\vspace{-0.2cm}
\subsection{Dataset Construction: Accounting Rules and Deterministic Validators}

Our benchmark curates explicit accounting principles specific to each statement type. We define four realistic accounting principle sets, each corresponding to the Balance Sheet, Cash Flow Statements, Statements of Equity, and Income Statements, respectively (\Cref{table:acc_rules}). Each rule set captures arithmetic consistency, hierarchical aggregation, cross-section dependencies, and conditional applicability relevant to the corresponding statement. The rule sets are non-overlapping and statement-specific rather than a single global rule pool. These principle/rule sets reflect the corresponding statements' distinct semantics and structure. While some rules require reasoning across multiple rows or sections within a record, FinRule-Bench does not enforce reconciliation constraints across different statement types. This design isolates rule-based reasoning within individual statement types and avoids confounding intra-statement reasoning with cross-statement consistency.

Each rule $r\in R$ is associated with a deterministic validator function $v_r(T)\in\{\texttt{True},\texttt{False}\}$. Validators implement the accounting logic programmatically and are used exclusively to generate ground-truth labels and to evaluate model predictions. Validator outputs are treated as the sole source of truth for model evaluation throughout all experiments.

\vspace{-0.2cm}
\subsection{Dataset Construction: Controlled Error Injection}

For controlled evaluation of the models, we construct error-injected variants of the base dataset by deterministically violating the predefined accounting principles. Error injection is guided by the accounting rule definitions per statement type, thereby ensuring realistic accounting errors rather than arbitrary perturbations which is done in some recent research papers \citep{choudhury-etal-2025-tabard, choudhury2026betterclausediscrepancybenchmark}. The clean financial statements and the controlled, rule-aware error-injected statements together represent a unified dataset, designed to evaluate the joint reasoning capabilities of the models.

The error-injection is carried out with respect to two task formulation-based regimes. For the single-violation regimes, exactly one accounting principle violation is injected per record. This constraint guarantees unambiguous rule-violation verification and identification, thereby isolating the intended reasoning requirement.
For the multi-violation regime, multiple accounting principle violations are injected within the same financial statement. These violations may affect different rows, sections, or rule types and require multi-step joint reasoning that mirrors realistic auditing scenarios. Across both regimes, error injection preserves the original table structure, schema, and formatting. Only numerical values or derived fields directly relevant to the targeted principles are modified. As a result, performance reflects reasoning over accounting rules rather than sensitivity to synthetic formatting changes.

\vspace{-0.2cm}
\subsection{Task Formulation over FinRule-Bench}

FinRule-Bench defines three auditing tasks that evaluate progressively stronger reasoning capabilities over structured financial tables. Each task corresponds to a distinct auditing problem and evaluates whether a model can interpret accounting principles, apply them consistently, and localize violations when present. These tasks correspond to different formulations over the same unified dataset and differ in violation cardinality and required model outputs.

\noindent\underline{\emph{Rule Verification.}}
Given a financial markdown table and a single accounting rule, the model predicts whether the table complies with the specified rule. This task evaluates basic rule compliance with a binary decision.

\noindent\underline{\emph{Rule Identification.}}
Given a financial table and a set of accounting rules, the model identifies the violated rule at a record level. Each record in the statement-specific table in this setting contains exactly one violated rule, ensuring unambiguous model evaluation and requiring discrimination among competing accounting principles.

\noindent\underline{\emph{Joint Rule Diagnosis.}}
Given a financial table and a set of accounting rules, the model first determines whether any violations are present per record per statement-type and, if so, identifies all violated rules with record-level attribution. This task evaluates multi-step diagnostic joint reasoning and coverage under realistic auditing conditions where multiple violations may co-occur.

All tasks operate within individual financial statements and do not assume explicit cross-statement consistency constraints. Formal task definitions are provided in Appendix~\ref{app:formal_tasks}.

\vspace{-0.2cm}
\subsection{Prompting and Inference Protocols}

We compare three prompting protocols that differ in structure and model supervision.

\noindent\textbf{Zero-shot.} Models recieve a task instruction, the serialized financial table $T$, and the relevant rule or rule set. No exemplars are provided in this protocol.

\noindent\textbf{Few-shot.} Models receive a small set of exemplars formatted in the same input/output style as the task, followed by the test instance. Exemplars are chosen to cover diverse rule types where possible and are fixed across the models evaluated to ensure fairness.

\noindent{\textbf{Few-shot with causal and counterfactual reasoning.} 
Beyond standard zero-shot and few-shot prompting, we introduce a causal-counterfactual prompting protocol to probe rule-based reasoning fidelity. We implement this prompting protocol using in-context learning, that vary systematically across three task formulations. Across all settings, causal and counterfactual reasoning structure is introduced exclusively through few-shot exemplars, rather than through architectural changes or additional supervision. In causal-counterfactual prompting, exemplars are constructed to make causal dependencies explicit rather than implicit. They illustrate both the condition that causes a rule violation and a minimal counterfactual modification that would remove the violation.

\noindent\underline{\emph{Rule Verification Task}}
In this setting, causal-counterfactual prompts are rule-specific and include compact exemplars that illustrate a single violation pattern together with a minimal counterfactual repair, such as correcting an arithmetic value or replacing a non-standard label. This evaluates whether exposure to causal and counterfactual structure improves the detection of clear violations without requiring discrimination among multiple rules.

\noindent\underline{\emph{Rule Identification Task}}
Here, causal-counterfactual exemplars are designed to support rule discrimination by clarifying which table attributes are causally relevant to each rule and the corresponding counterfactual diagnosis of other violated rules and apparent irregularities.

\noindent\underline{\emph{Joint Rule Diagnosis Task}}
In this joint diagnosis reasoning task, the exemplars contain detailed causal reasoning and counterfactual testing. However, the model’s output at inference time is restricted to a structured list of rule codes; explanations are not required. This setting evaluates whether models understand causal structure within each record in the table with respect to their rule compliance, sufficiently to achieve broad coverage when multiple violations interact under joint compositional reasoning.

Causal-counterfactual prompting does not enforce reasoning consistency and does not guarantee improved accuracy. Instead, it serves as a conditioning mechanism that exposes causal structure during in-context learning. Differences in performance across difficulty levels and prompting regimes therefore reflect the extent to which models internalize causal relevance and minimal interventions when reasoning about accounting rule violations, particularly in settings that require rule discrimination and multi-rule coverage. We emphasize that causal-counterfactual reasoning is introduced solely through exemplar design and does not rely on chain-of-thought prompting or post-hoc explanation scoring.

\vspace{-0.2cm}
\subsection{Model Evaluation and Verification Protocol}

Model evaluation is carried out using the deterministic rule validators corresponding to the defined accounting principles. The model outputs are parsed into structured representations and compared directly with the ground-truth. For rule verification and rule identification tasks, evaluation is based on exact-match protocol. For joint-rule diagnosis, the multi-step correctness requires accurate violation detection and accurate violated rule identification at record level. This protocol ensures clear objective fairness and reproducible evaluation.

With respect to the evaluation protocols, FinRule-Bench uses task-aligned, deterministic evaluation metrics computed via exact matching against rule validators. 
\noindent\underline{\emph{Rule Verification.}}
We report Accuracy, Precision, Recall, and F1-score to capture overall correctness and the trade-off between false alarms and missed violations.

\noindent\underline{\emph{Rule Identification.}}
We report Top-1 Accuracy as the primary metric. To account for rule imbalance, we additionally report Balanced Accuracy, Macro-F1, and Weighted-F1.

\noindent\underline{\emph{Joint Rule Diagnosis.}}
For violation detection, we report binary Accuracy and F1-score. For correctness, we report Exact Match and Micro-F1 to quantify partial coverage in multi-violation cases.

\noindent\underline{\emph{Efficiency Metrics.}}
For all tasks, we report Total Tokens, Average Tokens per Instance, Cost per Instance, and Total Cost to analyze performance–efficiency trade-offs.

\vspace{-0.3cm}
\subsection{Dataset Statistics}

We summarize the composition of FinRule-Bench across statement types and task formulations to provide a quantitative overview of the dataset. 
FinRule-Bench spans four financial statement types: Balance Sheets (BS), Cash Flow Statements (CF), Statements of Equity (SE), and Income Statements (SI). For each statement type, we define a statement-specific set of accounting rules and construct markdown tables with corresponding clean, single-violation, and multi-violation records, respectively. Each record in a table corresponds to a single company financial statement for a given fiscal period. Clean instances consist of fully compliant financial statements, while error-injected ones contain deterministically generated rule violations. For rule verification and rule identification, each instance contains at most one violation. For joint rule diagnosis, instances may contain multiple violations affecting different records or principles.
Across all statement types, the dataset covers a diverse range of accounting principles, including arithmetic consistency rules, structural and classification rules, conditional applicability rules, and multi-section reconciliation rules. This diversity ensures that evaluation supports a comprehensive assessment of financial rule-based reasoning. Table~\ref{tab:dataset_stats} and figure~\ref{fig:structural_complexity} summarize the dataset statistics, rule sets, and structural complexity distribution.

\begin{table}[t]
\centering
\footnotesize
\setlength{\tabcolsep}{3pt}
\caption{Dataset statistics for FinRule-Bench. Each record corresponds to a company's financial statement. Single-violation records contain exactly one violated rule, while multi-violation records contain multiple simultaneous violations.}
\label{tab:dataset_stats}
\begin{tabular}{@{}lcccc@{}}
\toprule
\textbf{Statement Type} & 
\textbf{\shortstack{\#\\Rules}} & 
\textbf{\shortstack{Ground-Truth\\Records}} & 
\textbf{\shortstack{Single-Violation\\Records}} & 
\textbf{\shortstack{Multi-Violation\\Records}} \\
\midrule
Balance Sheet (BS)       & 5  & 212  & 1060 & 100 \\
Cash Flow (CF)           & 3  & 318  & 954  & 100 \\
Statement of Equity (SE) & 3  & 287  & 871  & 100 \\
Income Statement (SI)    & 5  & 300  & 1500 & 100 \\
\midrule
\textbf{Total}           & 16 & 1117 & 4385 & 400 \\
\bottomrule
\end{tabular}
\end{table}

\begin{figure}[t]
\centering
\includegraphics[width=\columnwidth]{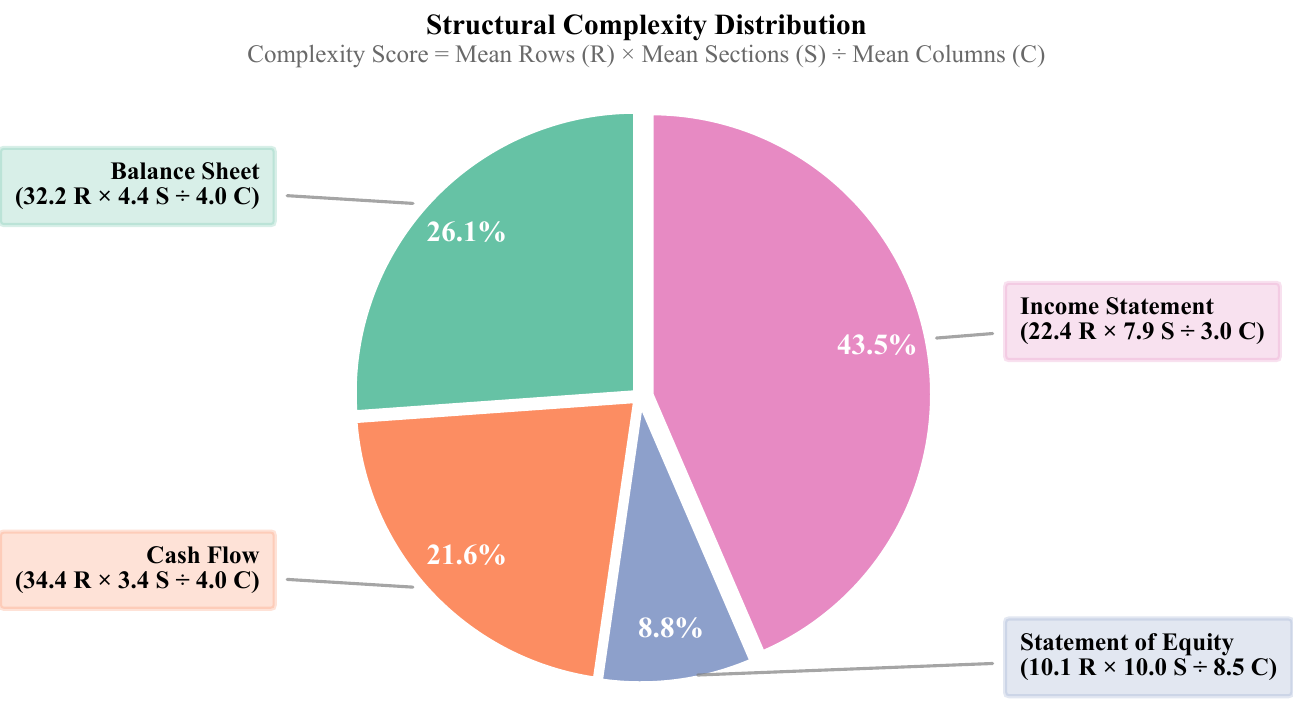}
\caption{Structural complexity distribution across financial statement types.}
\label{fig:structural_complexity}
\vspace{-0.2cm}
\end{figure}

\vspace{-0.2cm}
\subsection{Reproducibility and Artifacts}

All validators, error-injection prompts, exemplar prompts, and evaluation scripts are deterministic and included in the release artifacts. Model outputs are parsed by deterministic parsers to structured forms used for exact matching. We provide the rule definitions, validator code, exemplar prompts, and injection scripts to ensure complete reproducibility of reported results.

\section{Experiments}

\subsection{Experimental Setup}

We describe the models, data, prompting protocols, inference configuration, and evaluation procedures used in our experiments.

\noindent\textbf{Models.}
We evaluate four representative large language models: GPT-4o~\cite{openai2024gpt4ocard}, Gemini 2.5 Pro (thinking)~\cite{comanici2025gemini25pushingfrontier}, Gemini 2.0 Flash~\cite{hassabis2024gemini2}, and LLaMA 3.3~\cite{grattafiori2024llama3herdmodels} that are both reasoning-focused and efficiency-oriented architectures.

\noindent\underline{\emph{Data and splits.}}
The dataset used for all the experiments is described in Section~3. Clean markdown tables are validated prior to instantiation, and error-injected tables are used according to the targeted tasks (Section~3.4). Test sets are disjoint from any exemplars used for few-shot prompting. All per-task instance counts and split statistics are reported in Table~\ref{tab:dataset_stats}.

\noindent\underline{\emph{Prompting Protocols.}}
All models are evaluated under three prompt settings: zero-shot, few-shot, and few-shot with causal-counterfactual reasoning. Prompts include the task instruction, the serialized financial table $T$ in markdown format, and the relevant rule or rule set.

\noindent\underline{\emph{In-Context Exemplars.}}
In the two few-shot strategies, we provide a fixed number of labeled exemplars preceding the test instance. Exemplars follow the same input-output format as the target task and are selected to cover diverse rule types. The same exemplar set is used across all models to ensure fair comparison.

\noindent\underline{\emph{Decoding and Inference.}}
All models are evaluated using deterministic decoding with temperature set to zero and a set output length. No external tools, retrieval mechanisms, or post-processing heuristics are used beyond deterministic output parsing.

\noindent\underline{\emph{Evaluation Protocol.}}
Each test instance is evaluated independently. For tasks involving stochastic components in generation, results are averaged over multiple runs with identical settings. Ground-truth labels are generated exclusively by deterministic rule validators.

\noindent\underline{\emph{Cost and Token Accounting.}}
We record prompt tokens, completion tokens, and total tokens for every model invocation. Monetary cost is computed using publicly documented pricing for each model at the time of evaluation. The recorded token statistics enable performance-efficiency trade-off analyses.

\subsection{Experimental Results}
\subsubsection{Overall Performance Across Tasks.}
Table~\ref{tab:overall_performance} reports model performance of GPT-4o, Gemini 2.5 Pro (thinking), Gemini 2.0 Flash, and LLaMA 3.3 across rule verification, rule identification, and joint rule diagnosis on Finrule-Bench. These tasks progressively increase reasoning demands from isolated rule checking to multi-rule diagnostic reasoning.
On rule verification, all models achieve relatively high accuracy, indicating that basic compliance checking is largely within reach of modern LLMs. Performance differences among models are modest in this setting, suggesting that basic compliance checking against a single accounting principle is largely tractable for modern LLMs.
Performance drops substantially on rule identification, where models must isolate the violated principle among all accounting principles. Accuracy decreases across all models, and the performance gap between reasoning-oriented and efficiency-oriented models widens.
Performance degrades further on joint rule diagnosis. While Step-1 violation detection remains relatively high, Step-2 diagnostic exact match is low across all models. Even the strongest systems fail to localize all violated rules in most cases. However, this is where causal-counterfactual few-shot prompting proves to be effective, giving the largest relative gains across different models. These results highlight the challenges in exhaustive, record-level reasoning under multiple interacting constraints.
Also, we observe that the causal-counterfactual prompting yields clearer benefits for lightweight models like Gemini 2.0 Flash and LLaMA 3.3, where it consistently improves performance across verification, detection, and diagnostic localization. In contrast, reasoning-oriented models, such as GPT-4o and Gemini 2.5 Pro (thinking), exhibit inconsistent gains, and in some cases degraded performance, particularly on rule identification. This suggests that explicit causal scaffolding can introduce redundancy or interfere with discriminative decision-making in models that already perform extensive internal reasoning.
Overall, as task requirements progress from verification to discrimination and diagnosis, model performance degrades non-linearly, motivating the detailed diagnostic analyses presented in subsequent sections.

\begin{table}[t]
\centering
\footnotesize
\setlength{\tabcolsep}{1.5pt}
\caption{Overall Task performance on FinRule-Bench.}
\label{tab:overall_performance}
\resizebox{\linewidth}{!}{
\begin{tabular}{clcccc}
\toprule
\raisebox{1.5ex}{\textbf{Model}} 
& \raisebox{1.5ex}{\textbf{Baseline Prompt}} 
& \textbf{\shortstack[c]{Rule\\Verification}} 
& \textbf{\shortstack[c]{Rule\\Identification}} 
& \textbf{\shortstack[c]{Diagnosis\\Step-1}} 
& \textbf{\shortstack[c]{Diagnosis\\Step-2}} \\
\cmidrule(lr){3-6} 
& 
& \raisebox{1.5ex}{\textbf{Acc}} 
& \raisebox{1.5ex}{\textbf{Acc}} 
& \raisebox{1.5ex}{\textbf{Acc}} 
& \textbf{\shortstack[c]{Exact\\Match}} \\
\midrule

\multirow{4}{*}{GPT-4o}
& \textbf{Zero-shot}     
& 0.640 & 0.592 & 0.713 & 0.234 \\
& \textbf{Few-shot}      
& 0.736 & \textbf{0.689} & 0.786 & 0.297 \\
\cmidrule{2-6}
& \textbf{Few-shot + CR} 
& \textbf{0.738} & 0.661 & \textbf{0.801} & \textbf{0.342} \\
\midrule

\multirow{4}{*}{\shortstack[c]{Gemini \\ 2.5 Pro \\ (Thinking)}}
& \textbf{Zero-shot}     
& 0.614 & 0.547 & 0.695 & 0.216 \\
& \textbf{Few-shot}      
& 0.629 & \textbf{0.612} & 0.728 & 0.259 \\
\cmidrule{2-6}
& \textbf{Few-shot + CR} 
& \textbf{0.643} & 0.601 & \textbf{0.742} & \textbf{0.304} \\
\midrule

\multirow{4}{*}{\shortstack[c]{Gemini \\ 2.0 Flash}}
& \textbf{Zero-shot}     
& 0.564 & 0.524 & 0.631 & 0.152 \\
& \textbf{Few-shot}      
& 0.566 & 0.558 & 0.667 & 0.189 \\
\cmidrule{2-6}
& \textbf{Few-shot + CR} 
& \textbf{0.587} & \textbf{0.569} & \textbf{0.684} & \textbf{0.223} \\
\midrule

\multirow{4}{*}{LLaMA 3.3}

& \textbf{Zero-shot}     
& 0.553 & 0.496 & 0.619 & 0.147 \\
& \textbf{Few-shot}      
& \textbf{0.692} & 0.532 & 0.654 & 0.181 \\
\cmidrule{2-6}
& \textbf{Few-shot + CR} 
& 0.681 & \textbf{0.541} & \textbf{0.672} & \textbf{0.215} \\
\bottomrule
\end{tabular}
}
\end{table}
\vspace{-0.2cm}

\subsubsection{Rule Complexity and Error Analysis}

In this analysis, we evaluate model performance with respect to stratified rule complexity, error type, and distribution across the formulated tasks. This study isolates the semantic properties of rules that drive failures across tasks, beyond what is visible from aggregate accuracy.

\noindent\underline{\emph{Rule complexity analysis.}}
Tables~\ref{tab:rule_complexity_easy_all}, \ref{tab:rule_complexity_medium_all}, and \ref{tab:rule_complexity_hard_all} report performance stratified by rule category for rule verification, rule identification, and joint rule diagnosis tasks, respectively. Across all tasks, arithmetic rules are consistently the easiest, achieving high performance under zero-shot prompting and showing limited gains from additional reasoning structure. Structural rules exhibit intermediate difficulty and benefit most from few-shot and causal-counterfactual prompting strategies, reflecting improved schema and section-level interpretation through exemplars.
Conditional and multi-record rules are the most challenging across settings. Zero-shot performance is low, and although few-shot prompting improves results, accuracy remains substantially below that of simpler straight-forward rule types. We also observe that the Causal-counterfactual prompting yields its clearest benefits for multi-record constraints, particularly in the joint diagnosis setting, where reasoning must be composed across multiple records. However, these gains includes a trade off with increased computational cost.


\begin{table}[t]
\centering
\footnotesize
\setlength{\tabcolsep}{3.5pt}
\vspace{-0.1cm}
\caption{Rule Complexity Analysis on Rule Verification.}
\label{tab:rule_complexity_easy_all}
\begin{tabular}{l cc ccc}
\toprule
 &  &  & \multicolumn{3}{c}{\textbf{Acc}} \\
\cmidrule(lr){4-6}
\textbf{Category} & \textbf{\#Rules} & \textbf{\#Instances} & \textbf{Zero-shot} & \textbf{Few-shot} & \textbf{Few-shot + CR} \\
\midrule
Arithmetic    & 2 & 499  & 0.916 & 0.657 & 0.759 \\
Structural    & 5 & 424  & 0.735 & 0.794 & 0.791 \\
Conditional   & 6 & 1357 & 0.597 & 0.978 & 0.991 \\
Multi-record  & 3 & 510  & 0.916 & 0.822 & 0.790 \\
\bottomrule
\end{tabular}
\end{table}



\begin{table}[t]
\centering
\footnotesize
\setlength{\tabcolsep}{3.5pt}
\vspace{-0.1cm}
\caption{Rule Complexity Analysis for Rule Identification.}
\label{tab:rule_complexity_medium_all}
\begin{tabular}{l cc ccc}
\toprule
 &  &  & \multicolumn{3}{c}{\textbf{Accuracy}} \\
\cmidrule(lr){4-6}
\textbf{Rule Category} & \textbf{\#Rules} & \textbf{\#Instances} & \textbf{Zero-shot} & \textbf{Few-shot} & \textbf{Few-shot + CR} \\
\midrule
Arithmetic    & 2 & 499  & 0.922 & 0.910 & 0.896 \\
Structural    & 5 & 424  & 0.503 & 0.559 & 0.540 \\
Conditional   & 6 & 1357 & 0.590 & 0.703 & 0.590 \\
Multi-record  & 3 & 610  & 0.159 & 0.184 & 0.257 \\
\bottomrule
\end{tabular}
\vspace{2pt}
\end{table}

\begin{table}[t]
\centering
\footnotesize
\setlength{\tabcolsep}{4pt}
\vspace{-0.1cm}
\caption{Rule Complexity Analysis for Joint Rule Diagnosis.}
\label{tab:rule_complexity_hard_all}
\begin{tabular}{l cc cc}
\toprule
 &  &  & \multicolumn{2}{c}{\textbf{Metrics}} \\
\cmidrule(lr){4-5}
\textbf{Rule Category} & \textbf{\#Rules} & \textbf{\#Instances} & \textbf{Step-1 (Accuracy)} & \textbf{Step-2 (Micro-F1)} \\
\midrule
Arithmetic    & 2 & 79  & 0.797 & 0.545 \\
Structural    & 5 & 136 & 0.824 & 0.586 \\
Conditional   & 6 & 285 & 0.961 & 0.910 \\
Multi-record  & 3 & 144 & 0.576 & 0.591 \\
\bottomrule
\end{tabular}
\vspace{2pt}
\begin{tablenotes}
\small
\item \textit{Note:} Results are micro-averaged over instances within each rule category and aggregated across all financial statement types on the Model GPT-4o. Exact Match is not reported at the category level due to multi-rule aggregation.
\end{tablenotes}
\vspace{-0.3cm}
\end{table}

\noindent\underline{\emph{Error Type Analysis: Joint Rule Diagnosis.}}
Here, we analyze the failure modes on the joint rule diagnosis task. The results reveal a consistent gap between Step-1 violation detection and Step-2 diagnostic localization across all rule categories. While models frequently detect the presence of violations, they fail to identify all violated rules or to localize them correctly, especially for conditional and multi-record constraints.
Error-type analysis further shows that partial detection and incorrect localization errors are predominant across statement types. False positives are common for numerical constraints, while false negatives persist for context-dependent rules. These patterns indicate that joint diagnosis failures are driven primarily by incomplete coverage and misattribution rather than by an inability to detect anomalies.
Figure~\ref{fig:error_distribution} further illustrates the distribution of error types across statement types and prompting protocols for joint rule diagnosis. Across all settings, partial detection and mixed errors dominate, indicating that models frequently identify some violated rules but fail to produce complete or correctly localized diagnoses, confirming that joint diagnosis failures arise primarily from incomplete coverage and misattribution rather than from a failure to detect anomalies.
These results demonstrate that model failures in FinRule-Bench are structured and rule-dependent. Performance degrades systematically as tasks require reasoning over contextual applicability and cross-section dependencies, confirming that the benchmark isolates fundamental limitations in diagnostic reasoning.

\begin{figure}[t]
    \centering
    \includegraphics[width=\columnwidth]{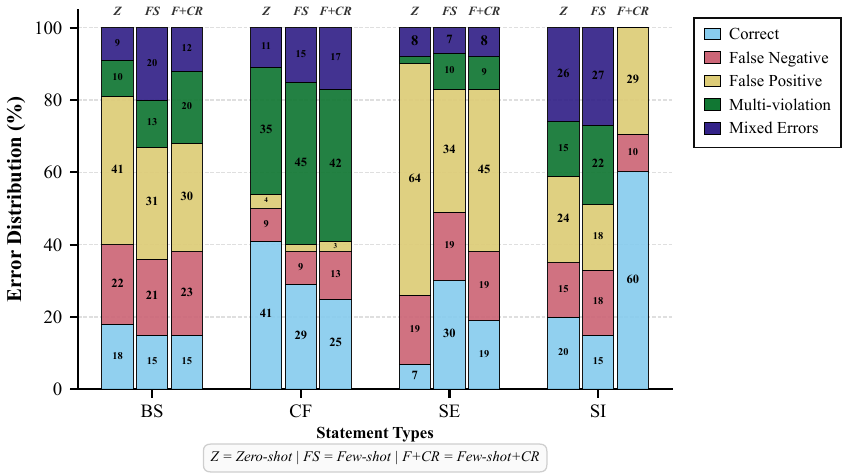}
    \vspace{-0.1cm}
    \caption{Error distribution across prompting methods.}
    \label{fig:error_distribution}
    \vspace{-0.3cm}
\end{figure}

\subsubsection{Performance-Cost Trade Offs}

We analyze the trade-off between performance and token consumption of all the prompting strategies across tasks to assess the cost-effectiveness of structured reasoning in FinRule-Bench.

\noindent\underline{\emph{Rule Verification.}}
Figure~\ref{fig:scalability_easy} shows that moving from zero-shot to few-shot prompting yields consistent accuracy gains at a modest increase in token usage. However, few-shot with causal-counterfactual reasoning substantially increases token consumption while providing limited statement-dependent accuracy improvements. For complex statements such as Balance Sheets and Cash Flow Statements, causal-counterfactual prompting can be cost-effective, whereas for simpler statements, it introduces token overhead without significant gains.

\noindent\underline{\emph{Rule Identification.}}
In this task, we observe that few-shot prompting improves identification accuracy relative to zero-shot across most statement types, but adding causal-counterfactual reasoning significantly increases token usage while failing to improve, and often degrading, discrimination performance~\ref{fig:scalability_easy}. This pattern holds across Balance Sheet, Statement of Equity, and Income Statement evaluations, indicating that verbose reasoning is poorly aligned with tasks that require identifying a single violated rule from a given rule set.

\noindent\underline{\emph{Joint Rule Diagnosis.}}
Figure~\ref{fig:scalability_hard} highlights a different trade-off. While causal-counterfactual prompting yields only modest gains for Step-1 violation detection, it provides larger relative improvements for Step-2 exact-match localization, particularly for statements with multiple interacting constraints. However, these gains come at a substantial token cost, with few-shot+CR increasing token usage relative to standard few-shot prompting. These results show that structured causal reasoning is most beneficial when exhaustive multi-rule localization is required. A consistent statement-level pattern also emerges. Causal-counterfactual prompting yields its strongest and most stable gains on Income Statement (SI) evaluations, which exhibit the highest structural complexity among the statement types in FinRule-Bench. SI statements contain dense interdependencies between subtotals, conditional line items, and cross-section aggregations, increasing the likelihood of interacting constraints~\ref{fig:structural_complexity}. In this setting, explicit causal exemplars and minimal counterfactual repairs more effectively guide models toward exhaustive localization of violated rules. It clearly shows its effectiveness scales with structural complexity rather than with task difficulty alone.

Overall, our performance efficiency analysis shows that structured prompting is not uniformly cost-effective. Additional reasoning overhead yields limited benefits for verification and discrimination tasks, but can improve on diagnostic localization tasks in joint rule diagnosis. This underscores the need for task-aware prompting strategies that balance accuracy against computational efficiency.

\begin{figure}[t]
\centering
\includegraphics[height=6.8cm, width=0.95\linewidth]{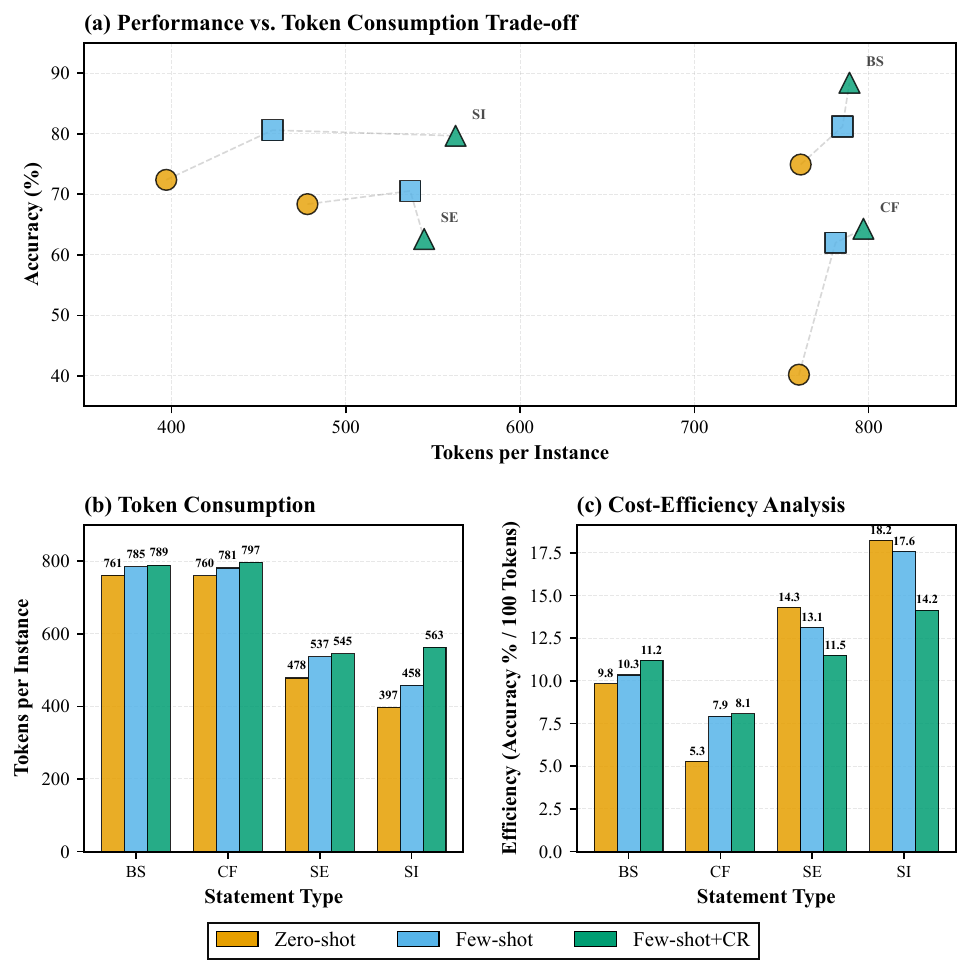}
\caption{Performance-cost analysis for rule verification.} 
\label{fig:scalability_easy}
\end{figure}

\begin{figure}[t]
\centering
\includegraphics[height=7cm, width=0.98\linewidth]{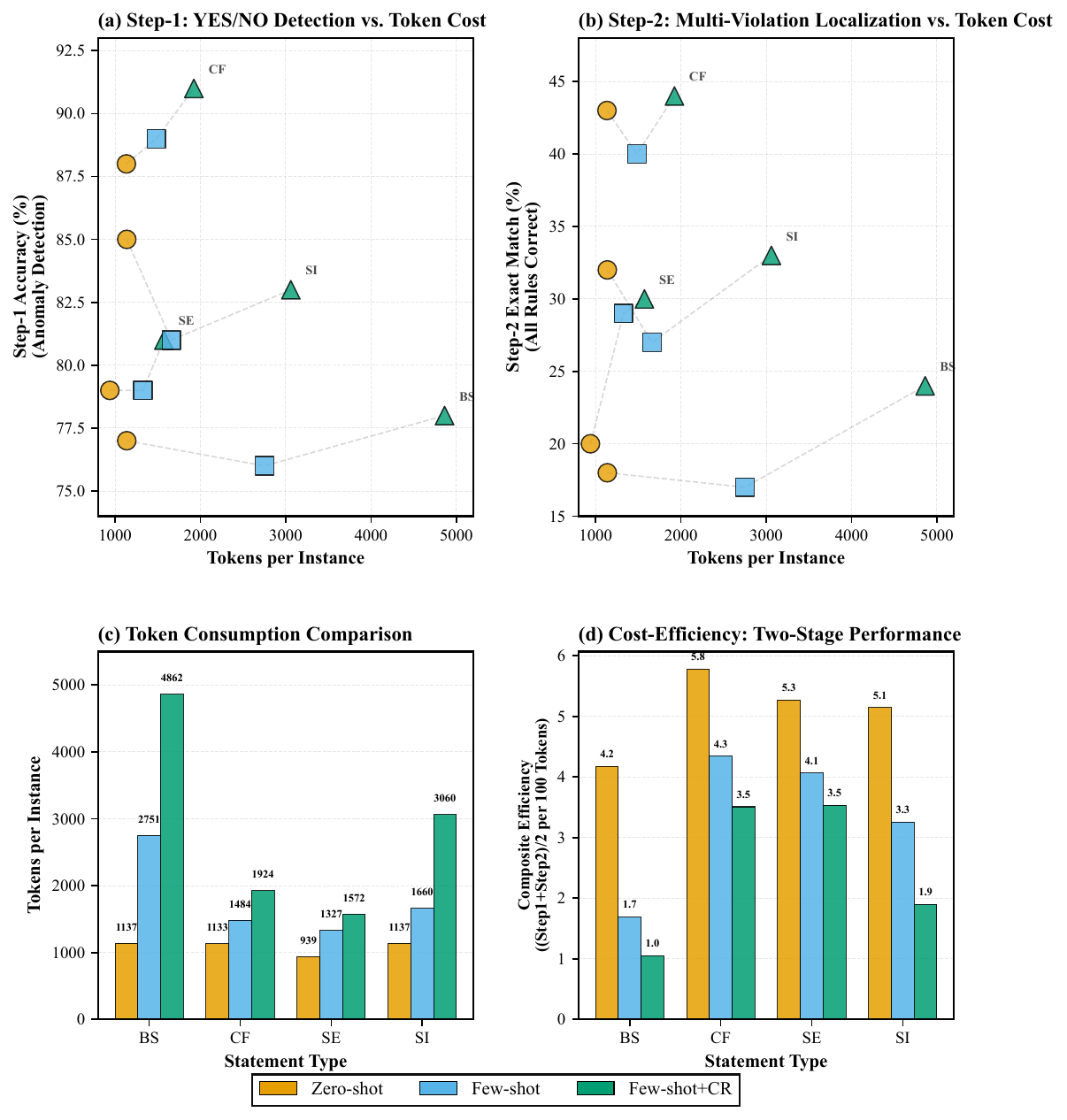}
\caption{Performance-cost analysis for the joint rule diagnosis task - Step-1 anomaly detection and Step-2 localization.}
\label{fig:scalability_hard}
\end{figure}

\begin{figure}[t]
\centering
\includegraphics[height=6.8cm, width=0.95\linewidth]{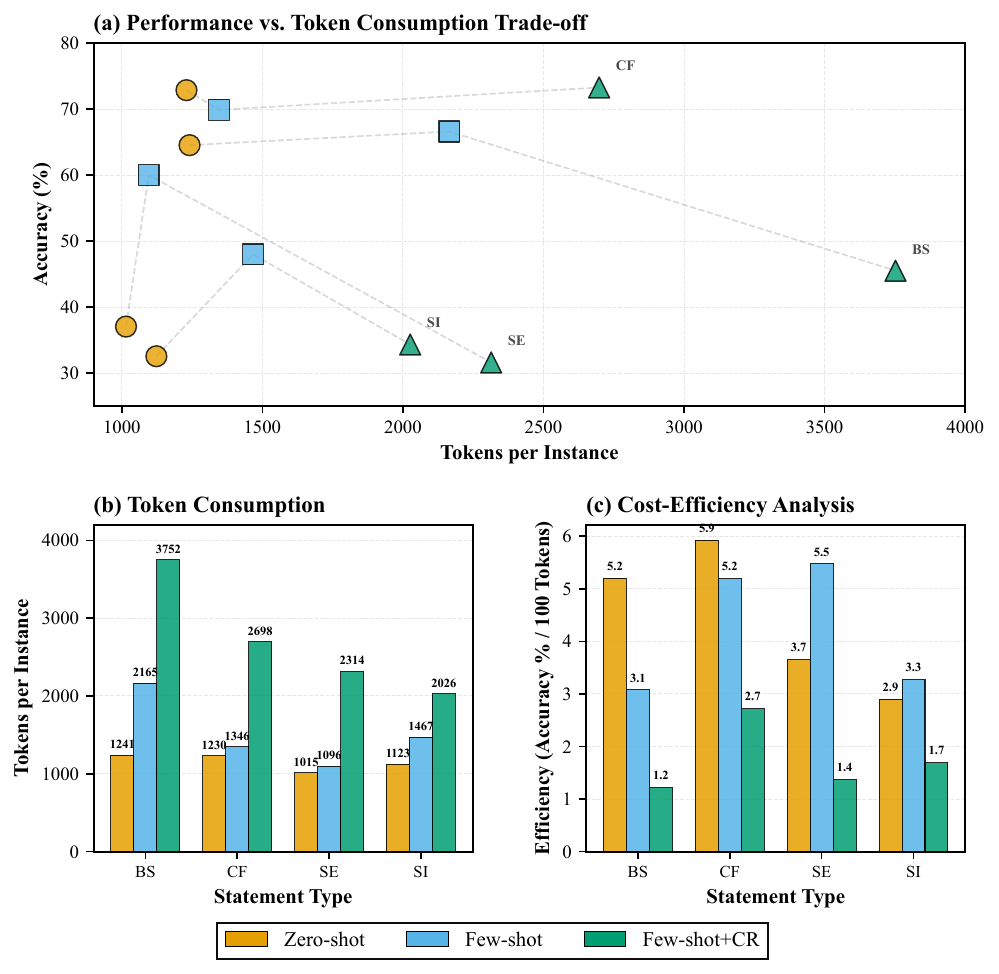} \caption{Performance-cost analysis for rule identification.}
\label{fig:scalability_medium}
\end{figure}

\section{Limitations}

FinRule-Bench evaluates rule-governed reasoning under the assumption that relevant accounting principles are provided and that financial tables are complete and internally coherent. The benchmark does not address rule discovery, reasoning under noisy or missing data, cross-statement reconciliation, or domain knowledge retrieval beyond the supplied rules. It is therefore intended as a focused diagnosis of joint-reasoning fidelity and not as a general-purpose financial analysis benchmark. We also do not study supervised fine-tuning or enforce chain-of-thought generation. Our goal is to evaluate the behavior of pretrained foundation models under fixed parameters and controlled task conditions, rather than to optimize performance through additional supervision or explanation constraints. While fine-tuning or explicit reasoning traces may improve accuracy, they also introduce confounding factors, optimization choices, and explanation verbosity that are outside the scope of our benchmark.

\section{Conclusion}

We introduced \textbf{FinRule-Bench}, a benchmark for evaluating rule-based compliance reasoning over financial tables under explicit accounting principles. Unlike prior financial and tabular benchmarks that emphasize question answering or anomaly detection on synthetically perturbed data, FinRule-Bench evaluates enforcement of formal accounting rules on ground-truth financial statements. The benchmark defines three auditing tasks i.e., single-rule verification, violated-rule identification, and joint multi-error diagnosis capturing progressively stronger reasoning requirements. Across models and prompting strategies, results show that while large language models reliably handle isolated arithmetic checks, performance degrades sharply for structural reasoning, conditional applicability, and composition across multiple interacting rules. Joint diagnosis in particular reveals a gap between detecting violations and achieving complete, record-level localization. These failures are systematic and rule-dependent, with performance declining as reasoning demands move from surface-level numerical consistency to exhaustive contextual and cross-record enforcement. By grounding evaluation in explicit accounting principles and realistic financial statements, FinRule-Bench provides a principled framework for assessing joint diagnostic reasoning in high-stakes tabular domains and motivates future work toward reliable end-to-end compliance verification.


\newpage
\section*{Ethics Statement}

This work introduces \textit{FinRule-Bench}, a research benchmark for evaluating the rule-based reasoning and diagnostic capabilities of large language models over structured financial statements. The benchmark is intended solely for \emph{research and evaluation purposes}, with the goal of analyzing reasoning fidelity, coverage, and failure modes under explicit accounting principles. It is not designed for deployment in real-world auditing, regulatory enforcement, or financial decision-making systems. All financial statements used in this benchmark are sourced from publicly available corporate disclosures, primarily Form 10-K filings released under regulatory requirements. The dataset operates exclusively on aggregated financial statement tables and accounting line items. No personally identifiable information, individual-level financial records, or confidential data are collected, inferred, or reconstructed. As such, the dataset does not introduce new privacy risks beyond those inherent in the original public filings.

To enable controlled evaluation, FinRule-Bench includes deterministically generated, rule-aware error-injected variants of otherwise correct financial statements. These synthetic violations are introduced solely to test model reasoning and diagnostic completeness and are explicitly labeled as evaluation artifacts. Performance on this benchmark should not be interpreted as evidence that a model is suitable for real-world auditing or compliance verification without qualified human oversight. We acknowledge that the benchmark reflects the institutional and regulatory context of its source data, including accounting standards and reporting practices commonly used in U.S.-centric financial disclosures. Consequently, models evaluated on FinRule-Bench may exhibit domain- or jurisdiction-specific behavior. These limitations are documented, and future extensions to additional accounting frameworks and jurisdictions are encouraged. Human involvement in dataset construction was limited to curating accounting principles, implementing and validating deterministic rule validators, and verifying task definitions. Large language models were used only in a restricted auxiliary capacity (e.g., assisting with table transcription or manuscript editing) and were not used to generate ground-truth labels. All evaluation labels are produced by deterministic validators to ensure reproducibility and objectivity.

Overall, this benchmark aims to support responsible data mining and AI research by enabling transparent analysis of reasoning failures in structured financial domains and by discouraging premature or unqualified deployment of language models in high-stakes financial contexts.

%

\bibliographystyle{ACM-Reference-Format}
\bibliography{main}

\appendix
\clearpage
\appendix
\section{Formal Task Definitions}
\label{app:formal_tasks}

This appendix provides formal definitions of the tasks used in FinRule-Bench. These definitions are included for completeness and reproducibility. The main paper presents task descriptions in evaluation-oriented terms.

\subsection{Preliminaries}

Let $T$ denote a structured financial table consisting of $m$ records $t_1,\dots,t_m$. Let $R=\{r_1,\dots,r_k\}$ denote the set of accounting rules applicable to $T$. Each rule $r \in R$ is associated with a deterministic validator function
\[
v_r(T) \in \{\texttt{True}, \texttt{False}\},
\]
which evaluates whether $T$ satisfies rule $r$. Validator outputs are treated as the ground truth for model evaluation.

\subsection{Rule Verification}

In the rule verification task, the model is given a financial table $T$ and a single accounting rule $r \in R$. The objective is to determine whether $T$ satisfies the specified rule.

Formally, the task is to produce a binary output
\[
y \in \{\texttt{True}, \texttt{False}\}
\]
such that
\[
y = v_r(T).
\]

This task corresponds to binary classification and evaluates a model’s ability to verify an explicit accounting principle on a structured financial table.

\subsection{Rule Identification}

In the rule identification task, the model is given a financial table $T$ and a set of accounting rules $R$. Exactly one rule $r^\star \in R$ is violated.

The objective is to output a rule identifier $\hat{r}$ such that
\[
\hat{r} = r^\star,
\]
where the constraint
\[
\left|\{r \in R \mid v_r(T) = \texttt{False}\}\right| = 1
\]
holds for all records across the statement types in this task.

Evaluation is done with exact-match criteria between the predicted rule identifier and the violated rule.

\subsection{Joint Rule Diagnosis}

In the joint rule diagnosis task, the model is given a financial table $T$ and a set of accounting rules $R$. An arbitrary subset of rules may be violated.

The task consists of two stages.

\paragraph{Violation Detection.}
The model produces a binary decision
\[
d \in \{\texttt{Yes}, \texttt{No}\},
\]
where
\[
d = \texttt{Yes} \iff \exists r \in R \text{ such that } v_r(T) = \texttt{False}.
\]

\paragraph{Violation Identification.}
If $d = \texttt{Yes}$, the model produces a mapping
\[
M : \{1,\dots,m\} \rightarrow 2^{R},
\]
where $M(i)$ denotes the set of accounting rules identified by the model as violated for record $t_i$.

For a prediction to be correct, both of the following conditions must hold:
\begin{enumerate}
    \item The violation detection decision $d$ matches the ground truth.
    \item The predicted rule sets $M(i)$ match the validator-defined violated rules at the record level.
\end{enumerate}

This task evaluates diagnostic reasoning under multiple simultaneous violations and requires both accurate detection and complete localization.

\section{Accounting Rules and Validator Specifications}
\label{app:validators}

Here, we provide the accounting rules used in FinRule-Bench and the deterministic validators employed to generate ground-truth labels. Validators provide an executable specification of each rule and error-representative algorithms are shown below.

\subsection{Validator Interface}

Each accounting rule $r$ is associated with a deterministic validator function:
\[
v_r : T \rightarrow \{\texttt{True}, \texttt{False}\},
\]
where $T$ is a structured financial table represented as an ordered collection of records with numeric values and hierarchical metadata. A validator returns \texttt{True} if the table satisfies the rule and \texttt{False} otherwise.

Ground-truth labels for all tasks are computed by evaluating the relevant validators on the original or error-injected tables.

\subsection{Representative Rule Categories}

We group accounting rules into four categories based on the dominant reasoning operation required for validation.

\paragraph{Arithmetic Rules.}
Arithmetic rules enforce numeric consistency relations. For example, a Balance Sheet must satisfy the accounting identity:
\[
\text{Total Assets} = \text{Total Liabilities} + \text{Total Equity}.
\]

\begin{algorithm}
\small
\caption{Arithmetic Validator (BS01)}
\label{alg:bs01}
\Fn{\textnormal{validate\_bs01}(table)}{
    $assets \leftarrow table[\text{"Total Assets"}]$\;
    $liabilities \leftarrow table[\text{"Total Liabilities"}]$\;
    $equity \leftarrow table[\text{"Total Equity"}]$\;
    \tcp{Check accounting identity}
    \Return $assets = liabilities + equity$\;
}
\end{algorithm}

\paragraph{Structural Rules.}
Structural rules constrain the presence, placement, or labeling of records. These rules do not require arithmetic computation but depend on correct classification within the table hierarchy.

\begin{algorithm}
\small
\caption{Structural Validator (SI03)}
\label{alg:si03}
\Fn{\textnormal{validate\_si03}(table)}{
    \tcp{Non-controlled interests should not in Liabilities}
    \Return $\text{"Non-controlled interests"} \in table.labels()$\;
}
\end{algorithm}

\paragraph{Conditional Rules.}
Conditional rules apply only when specific conditions are met. Validation requires checking both applicability and correctness.

\begin{algorithm}
\small
\caption{Conditional Validator (SI05)}
\label{alg:si05}
\Fn{\textnormal{validate\_si05}(table)}{
    \If{$table.contains(\text{"Interest Expense"})$}{
        \tcp{Must not be None if present}
        \Return $table.value(\text{"Interest Expense"}) \neq \text{None}$\;
    }
    \Return \True\;
}
\end{algorithm}

\paragraph{Multi-section Rules.}
Multi-section rules capture dependencies across multiple rows or time periods, such as rollforward or reconciliation constraints.

\begin{algorithm}
\small
\caption{Multi-section Validator (SE01)}
\label{alg:se01}
\Fn{\textnormal{validate\_se01}(table)}{
    \For{$year \in table.years()$}{
        $begin \leftarrow table.value(year, \text{"Beginning Balance"})$\;
        $change \leftarrow table.value(year, \text{"Net Change"})$\;
        $end \leftarrow table.value(year, \text{"Ending Balance"})$\;
        \tcp{Rollforward check}
        \If{$begin + change \neq end$}{
            \Return \False\;
        }
    }
    \Return \True\;
}
\end{algorithm}

\subsection{Error Injection via Validator Inversion}

Error-injected instances are generated by applying minimal edits that invert the outcome of a target validator while preserving table structure. For a rule $r$, we construct a modified table $T'$ such that:
\[
v_r(T) = \texttt{True} \quad \text{and} \quad v_r(T') = \texttt{False}.
\]

For arithmetic rules, this typically involves perturbing a single numeric value. For structural or conditional rules, this involves deleting, renaming, or relocating a single record. For multi-record rules, a single component in the dependency chain is modified. All injections are deterministic and reversible.

\subsection{Validator Composition for Joint Diagnosis}

For the joint rule diagnosis task, ground truth is computed by evaluating all validators on the same table:
\[
V(T) = \{ r \in R \mid v_r(T) = \texttt{False} \}.
\]

A model prediction is considered correct if and only if it identifies exactly the set $V(T)$ and associates each violated rule with the correct record(s). This strict criterion ensures that partial detection and mislocalization are treated as distinct failure modes.

\subsection{Reproducibility Guarantees}

Given the same input table, validators always produce identical outputs. This design guarantees that all reported results are fully reproducible and that any performance differences arise solely from model behavior rather than annotation noise or evaluator ambiguity.
The accounting rules are provided in the table~\ref{table:acc_rules}

\section{Error Injection and Dataset Instantiation}
\label{app:error_injection}

This appendix describes the procedures used to create task-specific instances from the base corpus of clean financial statements. All procedures are deterministic, reversible, and implemented in the released artifact set.

\subsection{Notation and setup}

Let $T$ be a clean base table and $R=\{r_1,\dots,r_k\}$ the rule set for $T$ (statement-specific). Let $v_r(T)$ be the deterministic validator for rule $r$, which for a clean table satisfies $v_r(T)=\texttt{True}$ for all $r\in R$. An injection targeting rule $r$ constructs a perturbed table $T'$ such that $v_r(T')=\texttt{False}$ while preserving structural properties of $T$.

In our dataset:
\begin{itemize}
  \item Balance Sheet (BS): 5 rules.
  \item Cash Flow (CF): 3 rules.
  \item Statement of Equity (SE): 3 rules.
  \item Income Statement (SI): 5 rules.
\end{itemize}

\subsection{Single-violation injection}

For rule verification and rule identification tasks, we generate single-violation instances as follows. The process is deterministic and rule-aware:

\begin{algorithm}
\small
\caption{Single-violation injection}
\label{alg:single_injection}
\Fn{\textnormal{inject\_single}(T, r)}{
    \tcp{Precondition: $v_r(T) == \textnormal{True}$}
    Identify the validator checks and the minimal dependent field(s) for rule $r$\;
    Determine a minimal edit (numeric perturbation, label rename, or row relocation) that will cause $v_r$ to return False\;
    Apply the edit to $T$ to obtain $T'$\;
    Record the edit script $\delta = \textnormal{diff}(T, T')$ for reversibility\;
    \Return $(T', \delta)$\;
}
\end{algorithm}

\subsection{Multi-violation construction (Hard task)}

For joint diagnosis task, we create multi-violation instances containing multiple simultaneously violated rules. These are constructed deterministically from clean tables as follows:

\begin{algorithm}
\small
\caption{Multi-violation construction}
\label{alg:multi_injection}
\Fn{\textnormal{inject\_multi}(T, $\{r_{a}, r_{b}, \dots\}$)}{
    \tcp{Precondition: $v_{r}(T) == \textnormal{True}$ for all $r$}
    Initialize $T_0 \leftarrow T$\;
    \For{each target rule $r_i$ in a predefined sequence}{
        Compute $(T_i, \delta_i) \leftarrow \textnormal{inject\_single}(T_{i-1}, r_i)$\;
        Verify $v_{r_j}(T_i)$ for all rules $r_j$; if injection unintentionally flips non-target validators, apply secondary minimal edits to restore only desired flips (keeping determinism)\;
    }
    Record combined edit script $\Delta = \{\delta_1,\delta_2,\dots\}$\;
    \Return $(T_{\text{final}}, \Delta)$\;
}
\end{algorithm}

\subsection{Dataset instantiation policy}

Task views are derived deterministically from the same base set of clean tables:

\begin{itemize}
  \item \textbf{Clean instances:} original extracted markdown tables with no edits.
  \item \textbf{Rule Verification / Rule-violation identification instances:} for each rule, generate single-violation instances deterministically; Also ensure uniqueness of the violated rule per instance.
  \item \textbf{Joint multi-step diagnosis instances:} produce multi-violation records as described above; in our experiments, we produced 100 multi-violation records per statement type.
\end{itemize}

This policy ensures that all task views are comparable, reproducible, and derived from the same underlying real-world disclosures.

\section{Financial Statement Samples and Structural Characteristics}
\label{app:samples}

This appendix provides representative examples of the financial statements used in FinRule-Bench. We include both original source documents and their extracted tabular representations to illustrate data provenance, extraction fidelity, and structural complexity. These examples are provided for transparency and are not exhaustive.

\subsection{Original Financial Statement Sources}

Figure~\ref{fig:financial_statement} shows representative financial statements snapshots as reported in corporate filings. These images correspond to the original source documents from which FinRule-Bench tables are extracted.

\begin{figure*}[ht]
    \centering
    \begin{subfigure}[b]{0.49\textwidth}
        \centering
        \includegraphics[width=\linewidth]{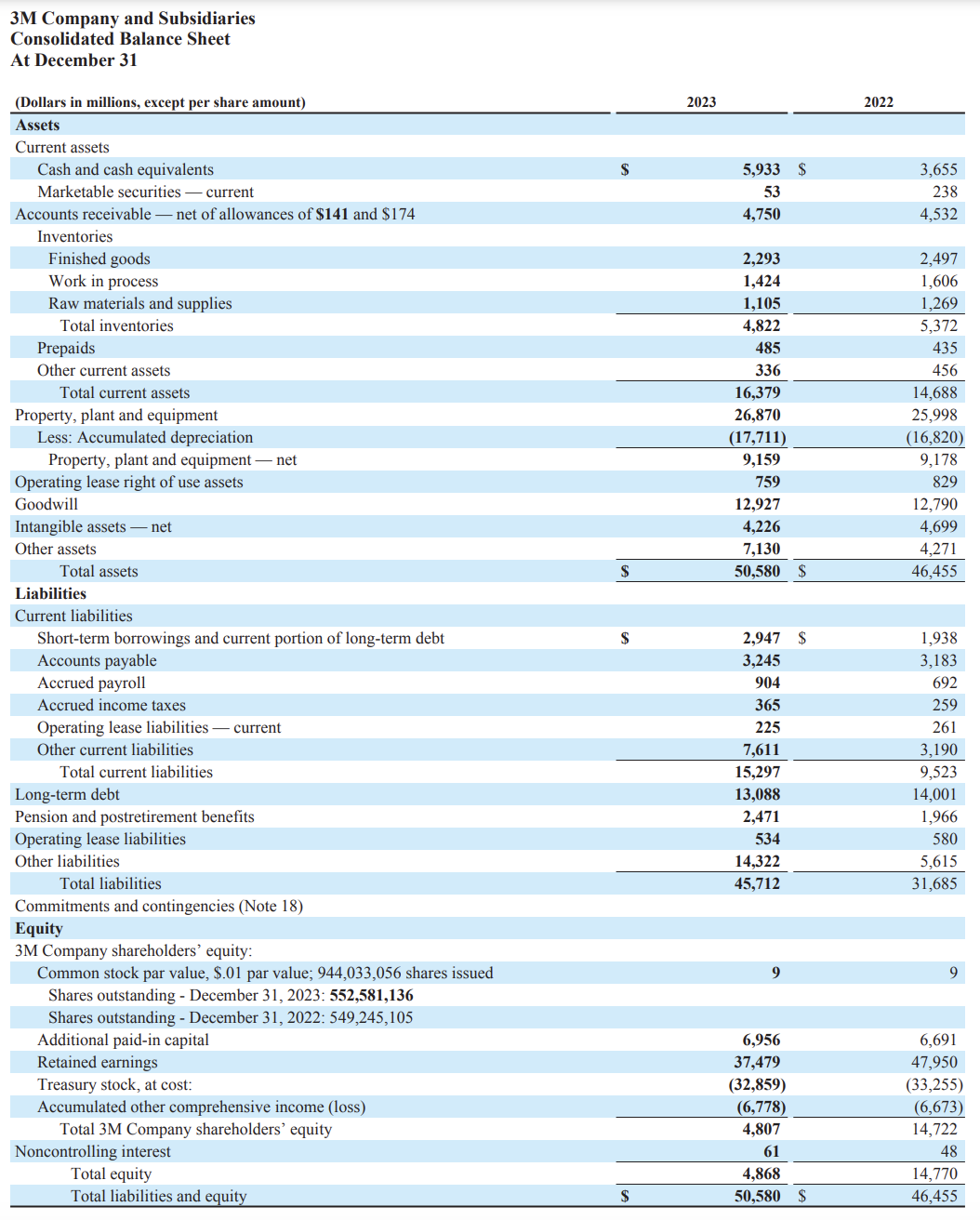}
        \caption{Balance Sheet}
        \label{fig:bs}
    \end{subfigure}
    \hfill
    \begin{subfigure}[b]{0.49\textwidth}
        \centering
        \includegraphics[width=\linewidth]{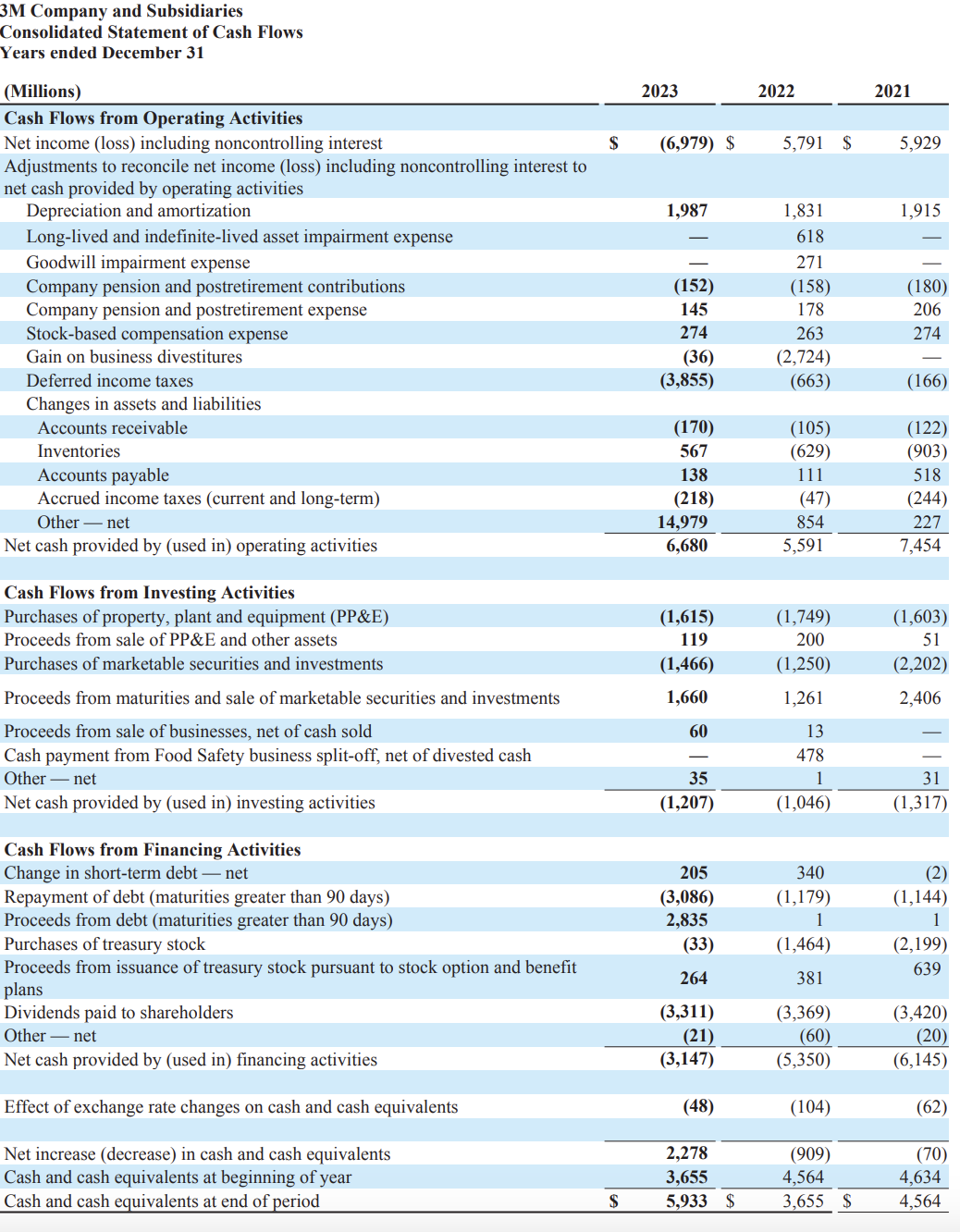}
        \caption{Statement of Cash Flow}
        \label{fig:cf}
    \end{subfigure}
    \vspace{5mm}
    \begin{subfigure}[b]{0.49\textwidth}
        \centering
        \includegraphics[width=\linewidth]{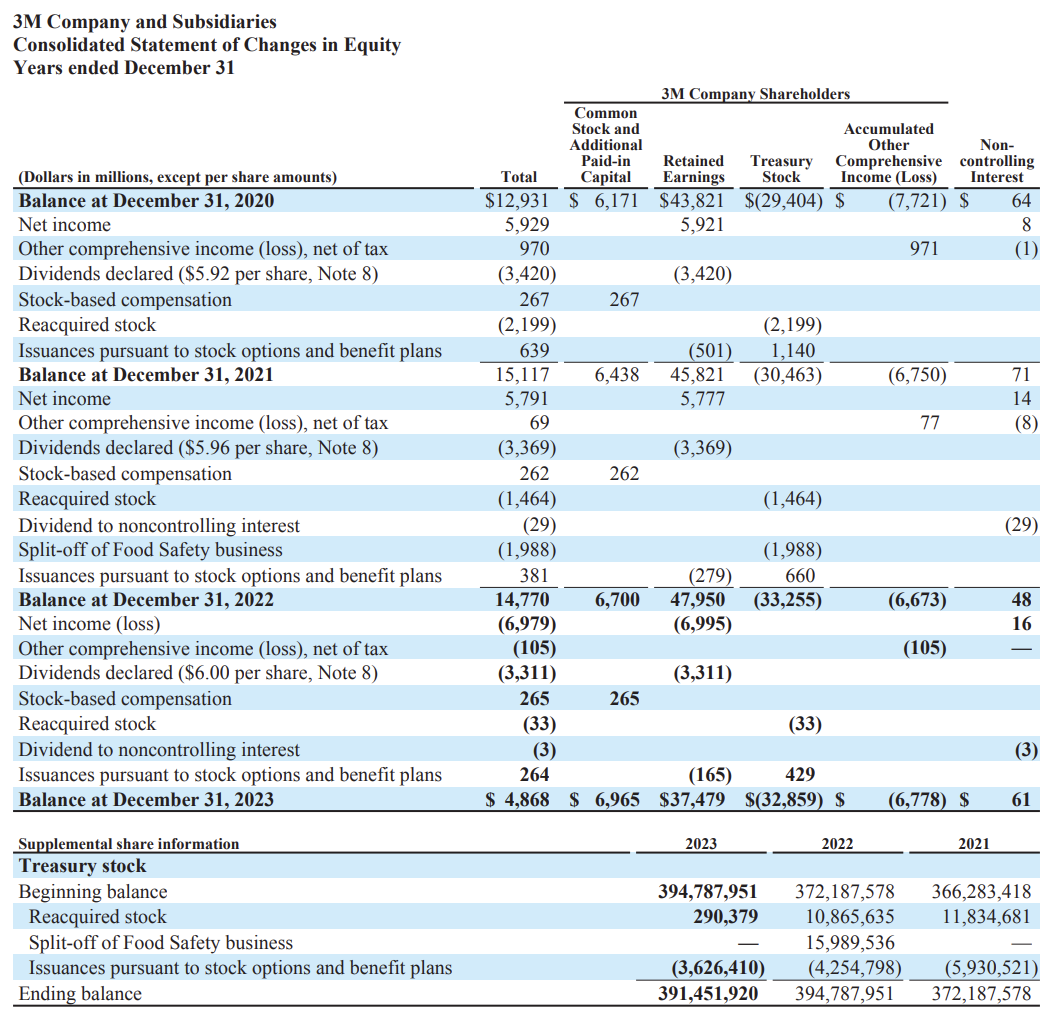}
        \caption{Statement of changes in Equity}
        \label{fig:se}
    \end{subfigure}
    \hfill
    \begin{subfigure}[b]{0.49\textwidth}
        \centering
        \includegraphics[width=\linewidth]{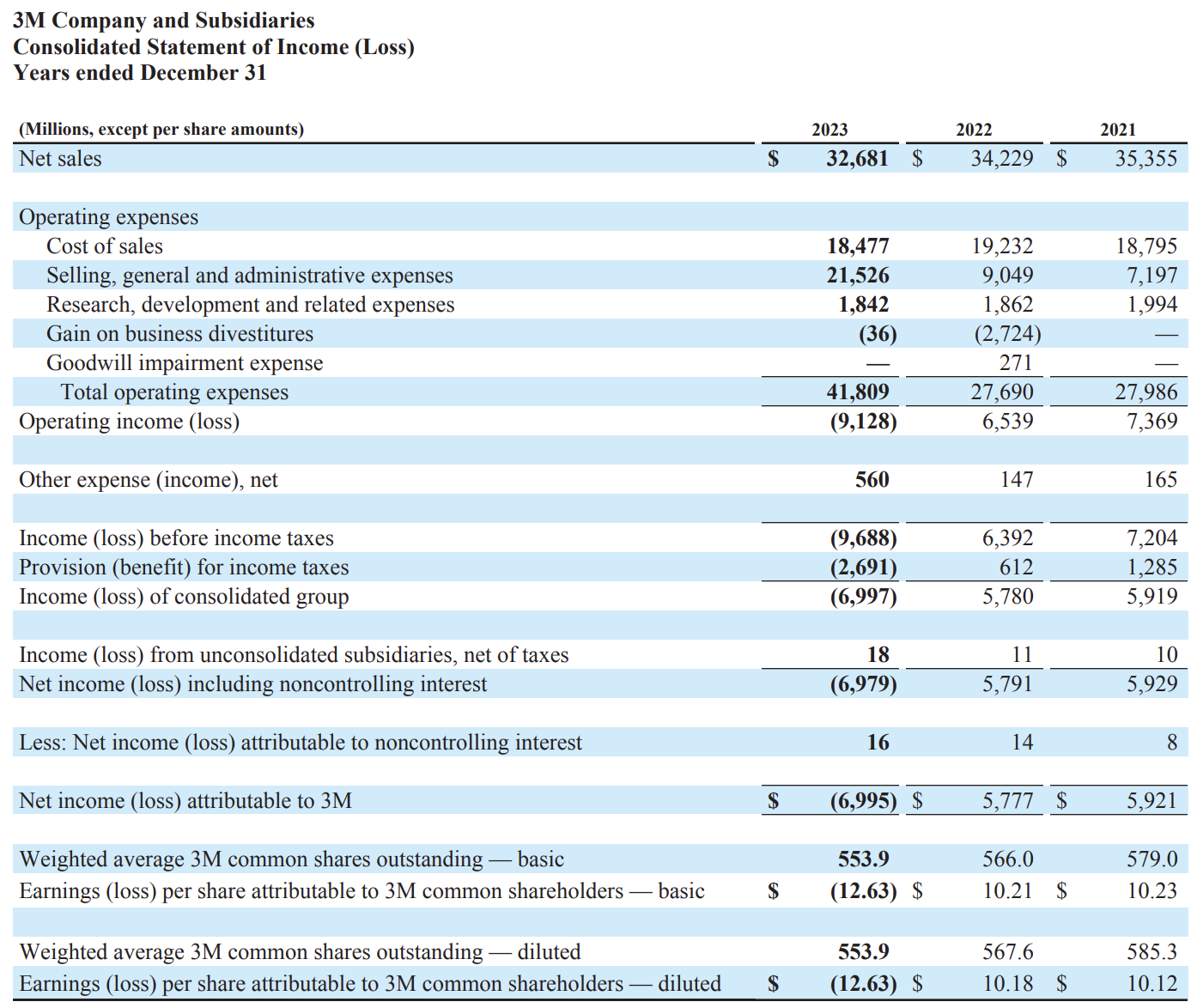}
        \caption{Income Statement}
        \label{fig:si}
    \end{subfigure}
    \caption{Raw Financial Statement screenshot images}
    \label{fig:financial_statement}
\end{figure*}

\subsection{Extracted Markdown Table Representation}

Table~\ref{table:table_samples} presents an example of a financial statement after extraction into the Markdown format used as model input. The extracted table preserves section hierarchy, record ordering, labels, and numerical values from the original filing snapshot~\ref{fig:financial_statement}.

\begin{table*}[ht]
\centering
\caption{Example extracted Balance Sheet (MasterCard, December 31, 2023). The table illustrates hierarchical structure, subtotals, cross-section totals, and sign conventions used in FinRule-Bench. Individual line items are omitted for brevity.}
\label{table:table_samples}
\small
\begin{tabular}{lllr}
\toprule
\textbf{Section} & \textbf{Subsection} & \textbf{Item} & \textbf{Value} \\
\midrule
Assets & Current Assets & Cash and cash equivalents & 8,588 \\
& & \dots & \dots \\
\cmidrule(lr){3-4}
& & \textbf{Total current assets} & \textbf{18,961} \\
\addlinespace
& Non-Current Assets & Goodwill & 7,660 \\
& & \dots & \dots \\
\midrule
\textbf{Total Assets} & & & \textbf{42,448} \\
\midrule
Liabilities & Current Liabilities & Accounts payable & 834 \\
& & \dots & \dots \\
\cmidrule(lr){3-4}
& & \textbf{Total current liabilities} & \textbf{16,264} \\
\addlinespace
& Non-Current Liabilities & Long-term debt & 14,344 \\
& & \dots & \dots \\
\midrule
\textbf{Total Liabilities} & & & \textbf{35,451} \\
\midrule
\textbf{Mezzanine Equity} & & Redeemable Non-controlling Interests & \textbf{22} \\
\midrule
Equity & Stockholders Equity & Class A treasury stock, at cost & -60,429 \\
& & \dots & \dots \\
\cmidrule(lr){3-4}
& & \textbf{Total Equity} & \textbf{6,975} \\
\midrule
\textbf{Total Liabilities and Equity} & & & \textbf{42,448} \\
\bottomrule
\end{tabular}
\end{table*}

\subsection{Structural Characteristics}

Financial statements in FinRule-Bench exhibit substantial variation in length, hierarchy, and internal dependencies. Tables range from compact statements with relatively flat structure to complex reports containing multiple sections, nested subsections, subtotal rows, and cross-record constraints.

\begin{table*}[t]
\centering
\footnotesize
\setlength{\tabcolsep}{3pt}
\caption{Structural complexity statistics.}
\label{tab:structural_stats}
\begin{tabular}{@{}lcccccccc@{}}
\toprule
\textbf{Statement Type} & 
\textbf{\shortstack{\#\\Records}} & 
\textbf{\shortstack{Avg\\Rows}} & 
\textbf{\shortstack{Min\\Rows}} & 
\textbf{\shortstack{Max\\Rows}} & 
\textbf{\shortstack{Avg\\Sections}} & 
\textbf{\shortstack{Frac.\\w/ Subsections}} & 
\textbf{\shortstack{Frac.\\w/ Subtotals}} & 
\textbf{\shortstack{Avg Cols\\(Min-Max)}} \\
\midrule
Balance Sheet (BS)       & 212 &  32.2 &   7 &  49 &  4.4 & 1.00 & 1.00 &  4.0 (4-4) \\
Cash Flow (CF)           & 318 &  34.4 &   7 &  60 &  3.4 & 1.00 & 1.00 &  4.0 (4-4) \\
Statement of Equity (SE) & 287 &  10.1 &   1 &  29 & 10.0 & 1.00 & 0.98 &  8.5 (5-16) \\
Income Statement (SI)    & 300 &  22.4 &   2 &  40 &  7.9 & 1.00 & 0.99 &  3.0 (3-3) \\
\bottomrule
\end{tabular}
\end{table*}

\section{Dataset Verification}
\label{app:human_verification}

In this section, we report the inter-annotator agreement metrics used to assess annotation consistency and reliability. We also describe the annotation guidelines provided to the two annotators, which define the verification criteria and procedures followed during manual review.

\subsection{Human Verification}

To assess the reliability of the dataset and the correctness of the controlled rule violations, we conducted a manual human verification on a randomly sampled 15\% subset of the perturbed financial tables. This subset was partitioned into two disjoint samples, denoted as $\alpha_1$ and $\alpha_2$, each covering 7.5\% of the dataset, to independently evaluate annotation consistency. Results are shown in Table~\ref{tab:human-verification}.

\begin{table}[H]
\centering
\caption{Inter-annotator agreement results for human verification}
\label{tab:human-verification}
\begin{tabular}{lcc}
\toprule
\textbf{Sample} & \textbf{Cohen’s Kappa (\%)} & \textbf{Jaccard Coefficient (\%)} \\
\midrule
$\alpha_1$ & 93.3 & 97.5 \\
$\alpha_2$ & 95.3 & 95.8 \\
\bottomrule
\end{tabular}
\end{table}

Two annotators with expertise in natural language processing and prior experience in the financial domain independently reviewed the sampled tables. Inter-annotator agreement was quantified using Cohen’s kappa to measure consistency in violated-rule identification and Jaccard’s coefficient to measure overlap in record-level violation localization. Agreement was computed separately for $\alpha_1$ and $\alpha_2$ to assess stability across independent samples. These agreement metrics serve as a validation of the dataset construction process and provide an additional check on the reliability of the deterministic rule validators used for large-scale annotation.

\subsection{Annotation Guidelines}

Annotators were provided with a detailed set of written guidelines prior to beginning the manual verification process to ensure consistency, clarity, and reproducibility across annotations. The guidelines were designed to standardize how accounting rule violations were interpreted, validated, and localized within structured financial tables.

Each annotator was instructed to independently review a perturbed financial table alongside its associated contextual information, including the stated accounting principle and the intended violation introduced through controlled error injection. Annotators were asked to evaluate whether the injected violation was \emph{(i)} correctly instantiated according to the specified accounting rule, \emph{(ii)} contextually justified given the structure and semantics of the financial statement, and \emph{(iii)} accurately localized to the relevant rows or records in the table. Annotators were explicitly instructed not to infer additional violations beyond those introduced, and to focus solely on verifying the correctness and scope of the provided violation.

The guidelines emphasized that annotations should be grounded in the explicit accounting principles supplied with each instance, rather than relying on implicit assumptions, external knowledge, or alternative accounting interpretations. In cases where a violation appeared ambiguous, annotators were instructed to mark their judgment based on whether the violation could be reasonably supported under the stated rule definition, without attempting to resolve ambiguity through subjective interpretation.

Annotators were also provided with clarifying examples illustrating common rule categories, including arithmetic consistency rules, structural classification rules, conditional applicability rules, and multi-record constraints. These examples were intended to align annotator judgments with the deterministic rule validators used in dataset construction, while preserving independent human assessment.

Participation in the annotation process was entirely voluntary. Both annotators are NLP researchers with prior experience in financial-domain data and accounting-related tasks, and they chose to contribute due to their interest in the research problem and its potential impact on evaluating reasoning reliability in high-stakes financial settings. Annotators were not exposed to any sensitive or personally identifiable information and were not asked to perform tasks beyond verification of synthetic, research-oriented data artifacts.

No adjudication or consensus-building was performed during annotation. Disagreements were retained and used exclusively for computing inter-annotator agreement metrics, which serve as an empirical measure of annotation consistency rather than as a mechanism for correcting labels. This design ensures that reported agreement reflects genuine independent judgment rather than post-hoc alignment.
\newpage
\section{Experimental Results}
\label{app:additional_results}

This section presents additional experimental results that complement the main findings in Section~4.

\begin{table}[h]
\centering
\footnotesize
\setlength{\tabcolsep}{3pt}
\begin{threeparttable}
\caption{Performance on Rule Compliance Verification (GPT-4o)}
\label{tab:easy_results_gpt4o}
\begin{tabular}{llccccc}
\toprule
\textbf{Statement} & \textbf{Method} & \textbf{Acc} & \textbf{Precision} & \textbf{Recall} & \textbf{F1-Score} & \textbf{Tokens (K)} \\
\midrule

\multirow{3}{*}{\textbf{BS}} 
& \textbf{Zero-shot}     & 0.749 & 0.824 & 0.890 & 0.855 & 967.7 \\
& \textbf{Few-shot}      & 0.812 & 0.882 & 0.894 & 0.888 & 998.0 \\
\cmidrule(lr){2-7}
& \textbf{Few-shot + CR} & \textbf{0.884} & \textbf{0.962} & \textbf{0.897} & \textbf{0.928} & 1003.1 \\
\midrule

\multirow{3}{*}{\textbf{CF}} 
& \textbf{Zero-shot}     & 0.402 & 0.696 & 0.361 & 0.476 & 982.5 \\
& \textbf{Few-shot}      & 0.620 & 0.819 & 0.634 & 0.714 & 1009.2 \\
\cmidrule(lr){2-7}
& \textbf{Few-shot + CR} & \textbf{0.643} & \textbf{0.849} & \textbf{0.638} & \textbf{0.728} & 1030.8 \\
\midrule

\multirow{3}{*}{\textbf{SE}} 
& \textbf{Zero-shot}     & 0.684 & 0.734 & \textbf{0.908} & \textbf{0.812} & 548.3 \\
& \textbf{Few-shot}      & \textbf{0.706} & \textbf{0.887} & 0.696 & 0.780 & 616.6 \\
\cmidrule(lr){2-7}
& \textbf{Few-shot + CR} & 0.626 & 0.753 & 0.746 & 0.750 & 616.1 \\
\midrule

\multirow{3}{*}{\textbf{SI}} 
& \textbf{Zero-shot}     & 0.724 & 0.834 & 0.835 & 0.834 & 713.9 \\
& \textbf{Few-shot}      & \textbf{0.806} & \textbf{0.882} & \textbf{0.885} & \textbf{0.884} & 825.3 \\
\cmidrule(lr){2-7}
& \textbf{Few-shot + CR} & 0.797 & 0.881 & 0.874 & 0.878 & 1014.2 \\
\midrule

\multirow{3}{*}{\textbf{Overall}} 
& \textbf{Zero-shot}     & 0.640 & 0.772 & 0.749 & 0.744 & 3212.4 \\
& \textbf{Few-shot}      & 0.736 & \textbf{0.868} & 0.777 & 0.816 & 3449.1 \\
\cmidrule(lr){2-7}
& \textbf{Few-shot + CR} & \textbf{0.738} & 0.861 & \textbf{0.789} & \textbf{0.821} & 3664.2 \\
\bottomrule
\end{tabular}
\end{threeparttable}
\end{table}

\begin{table}[h]
\centering
\footnotesize
\begin{threeparttable}
\caption{Performance on Rule Compliance Verification (Gemini 2.5 Pro)}
\label{tab:easy_results_gemini}
\setlength{\tabcolsep}{3pt}
\begin{tabular}{llccccc}
\toprule
\textbf{Statement} & \textbf{Method} & \textbf{Acc} & \textbf{Precision} & \textbf{Recall} & \textbf{F1-Score} & \textbf{Tokens (K)}  \\
\midrule

\multirow{3}{*}{\textbf{BS}} 

& \textbf{Zero-shot}     & \textbf{0.739} & \textbf{0.835} & \textbf{0.857} & \textbf{0.845} & 1272.0  \\
& \textbf{Few-shot}      & 0.676 & 0.810 & 0.798 & 0.804 & 2617.9  \\
\cmidrule{2-7}
& \textbf{Few-shot + CR} & 0.473 & 0.768 & 0.526 & 0.624 & 3316.7  \\
\midrule

\multirow{3}{*}{\textbf{CF}} 

& \textbf{Zero-shot}     & 0.619 & 0.803 & 0.652 & 0.720 & 2067.7  \\
& \textbf{Few-shot}      & 0.606 & 0.800 & \textbf{0.707} & \textbf{0.751} & 1595.3  \\
\cmidrule{2-7}
& \textbf{Few-shot + CR} & \textbf{0.628} & \textbf{0.886} & 0.578 & 0.700 & 2415.5  \\
\midrule

\multirow{3}{*}{\textbf{SE}} 

& \textbf{Zero-shot}     & 0.634 & 0.833 & 0.641 & 0.724 & 2114.6  \\
& \textbf{Few-shot}      & 0.585 & 0.795 & 0.602 & 0.685 & 2411.0  \\
\cmidrule{2-7}
& \textbf{Few-shot + CR} & \textbf{0.651} & \textbf{0.837} & \textbf{0.663} & \textbf{0.740} & 2820.8  \\
\midrule

\multirow{3}{*}{\textbf{SI}} 

& \textbf{Zero-shot}     & 0.459 & 0.757 & 0.517 & 0.615 & 2504.3  \\
& \textbf{Few-shot}      & 0.647 & 0.806 & \textbf{0.759} & 0.782 & 2637.8  \\
\cmidrule{2-7}
& \textbf{Few-shot + CR} & \textbf{0.658} & \textbf{0.833} & 0.738 & \textbf{0.783} & 3491.9  \\
\midrule

\multirow{3}{*}{\textbf{Overall}} 

& \textbf{Zero-shot}     & 0.614 & 0.792 & 0.615 & 0.726 & 7958.6  \\
& \textbf{Few-shot}      & 0.629 & 0.803 & \textbf{0.717} & \textbf{0.756} & 9262.0  \\
\cmidrule{2-7}
& \textbf{Few-shot + CR} & \textbf{0.643} & \textbf{0.832} & 0.676 & 0.746 & 12044.9  \\
\bottomrule
\end{tabular}
\end{threeparttable}
\end{table}

\begin{table}[h]
\centering
\footnotesize
\begin{threeparttable}
\caption{Performance on Rule Compliance Verification (Gemini 2.0)}
\label{tab:easy_results_gemini_overall}
\setlength{\tabcolsep}{3pt}
\begin{tabular}{llccccc}
\toprule
\textbf{Statement} & \textbf{Method} & \textbf{Acc} & \textbf{Precision} & \textbf{Recall} & \textbf{F1-Score} & \textbf{Tokens (K)}  \\
\midrule

\multirow{3}{*}{\textbf{BS}} 

& \textbf{Zero-shot}     & \textbf{0.759} & 0.870 & 0.836 & 0.853 & 1109.4  \\
& \textbf{Few-shot}      & 0.519 & \textbf{0.900} & \textbf{0.862} & \textbf{0.881} & 1142.3  \\
\cmidrule{2-7}
& \textbf{Few-shot + CR} & 0.297 & 0.840 & 0.739 & 0.786 & 2257.7  \\
\midrule

\multirow{3}{*}{\textbf{CF}} 

& \textbf{Zero-shot}     & 0.805 & 0.803 & \textbf{0.652} & 0.417 & 1045.6  \\
& \textbf{Few-shot}      & 0.858 & 0.919 & 0.540 & 0.680 & 1070.7 \\
\cmidrule{2-7}
& \textbf{Few-shot + CR} & \textbf{0.907} & \textbf{0.953} & 0.632 & \textbf{0.760} & 2522.6  \\
\midrule

\multirow{3}{*}{\textbf{SE}} 

& \textbf{Zero-shot}     & 0.105 & 0.705 & 0.714 & 0.710 & 639.8  \\
& \textbf{Few-shot}      & 0.150 & 0.717 & 0.719 & 0.718 & 761.8  \\
\cmidrule{2-7}
& \textbf{Few-shot + CR} & \textbf{0.190} & \textbf{0.815} & \textbf{0.752} & \textbf{0.782} & 819.9  \\
\midrule

\multirow{3}{*}{\textbf{SI}} 

& \textbf{Zero-shot}     & 0.130 & 0.792 & 0.661 & 0.720 & 791.2  \\
& \textbf{Few-shot}      & \textbf{0.210} & \textbf{0.843} & \textbf{0.851} & \textbf{0.847} & 906.0  \\
\cmidrule{2-7}
& \textbf{Few-shot + CR} & 0.087 & 0.808 & 0.771 & 0.789 & 3769.9 \\
\midrule

\multirow{3}{*}{\textbf{Overall}} 

& \textbf{Zero-shot}     & \textbf{0.450} & 0.793 & 0.716 & 0.675 & 3586.1  \\
& \textbf{Few-shot}      & 0.434 & 0.845 & \textbf{0.743} & \textbf{0.781} & 3880.8  \\
\cmidrule{2-7}
& \textbf{Few-shot + CR} & 0.370 & \textbf{0.854} & 0.724 & 0.779 & 9370.1 \\
\bottomrule
\end{tabular}
\end{threeparttable}
\end{table}

\begin{table}[h]
\centering
\footnotesize
\begin{threeparttable}
\caption{Performance on Rule Compliance Verification (LLaMA)}
\label{tab:easy_results_llama_overall}
\setlength{\tabcolsep}{3pt}
\begin{tabular}{llccccc}
\toprule
\textbf{Statement} & \textbf{Method} & \textbf{Acc} & \textbf{Precision} & \textbf{Recall} & \textbf{F1-Score} & \textbf{Tokens (K)} \\
\midrule

\multirow{3}{*}{\textbf{BS}} 

& \textbf{Zero-shot}     & 0.606 & 0.801 & 0.709 & 0.748 & 1006.1  \\
& \textbf{Few-shot}      & 0.749 & 0.873 & \textbf{0.818} & \textbf{0.845} & 1036.6  \\
\cmidrule{2-7}
& \textbf{Few-shot + CR} & \textbf{0.759} & \textbf{0.959} & 0.743 & 0.837 & 2016.7  \\
\midrule

\multirow{6}{*}{\textbf{CF}} 

& \textbf{Zero-shot}     & 0.458 & \textbf{0.808} & 0.364 & 0.502 & 1027.8  \\
& \textbf{Few-shot}      & 0.567 & 0.765 & 0.609 & 0.678 & 1053.9  \\
\cmidrule{2-7}
& \textbf{Few-shot + CR} & \textbf{0.579} & 0.762 & \textbf{0.638} & \textbf{0.695} & 2435.7  \\
\midrule

\multirow{3}{*}{\textbf{SE}} 

& \textbf{Zero-shot}     & 0.628 & 0.806 & 0.663 & 0.728 & 584.3  \\
& \textbf{Few-shot}      & \textbf{0.745} & \textbf{0.915} & \textbf{0.727} & \textbf{0.810} & 653.2  \\
\cmidrule{2-7}
& \textbf{Few-shot + CR} & 0.688 & 0.899 & 0.657 & 0.760 & 692.1 \\
\midrule

\multirow{3}{*}{\textbf{SI}} 

& \textbf{Zero-shot}     & 0.520 & 0.791 & 0.577 & 0.667 & 770.2  \\
& \textbf{Few-shot}      & \textbf{0.706} & 0.839 & \textbf{0.801} & \textbf{0.819} & 883.3  \\
\cmidrule{2-7}
& \textbf{Few-shot + CR} & 0.698 & \textbf{0.851} & 0.774 & 0.811 & 3376.8 \\
\midrule

\multirow{3}{*}{\textbf{Overall}} 

& \textbf{Zero-shot}     & 0.553 & 0.802 & 0.578 & 0.661 & 3388.4  \\
& \textbf{Few-shot}      & \textbf{0.692} & 0.848 & \textbf{0.739} & \textbf{0.788} & 3627.0  \\
\cmidrule{2-7}
& \textbf{Few-shot + CR} & 0.681 & \textbf{0.868} & 0.703 & 0.776 & 8521.3  \\
\bottomrule
\end{tabular}
\end{threeparttable}
\end{table}

\begin{table}[ht]
\centering
\footnotesize
\begin{threeparttable}
\caption{Performance on Single Violated Rule Identification (GPT-4o)}
\label{tab:medium_results}
\setlength{\tabcolsep}{2pt}
\begin{tabular}{llcccccc}
\toprule
\textbf{Statement} & \textbf{Method} & \textbf{Acc} & \textbf{Bal. Acc} & \textbf{M-Prec} & \textbf{M-Rec} & \textbf{M-F1} & \textbf{W-F1} \\
\midrule

\multirow{3}{*}{\textbf{BS}} 

& \textbf{Zero-shot}     & 0.663 & 0.663 & \textbf{0.751} & 0.663 & 0.641 & 0.641 \\
& \textbf{Few-shot}      & \textbf{0.724} & \textbf{0.724} & 0.750 & \textbf{0.724} & \textbf{0.702} & \textbf{0.702} \\
\cmidrule{2-8}
& \textbf{Few-shot + CR} & \textbf{0.724} & \textbf{0.724} & 0.750 & \textbf{0.724} & \textbf{0.702} & \textbf{0.702} \\
\midrule

\multirow{3}{*}{\textbf{CF}} 

& \textbf{Zero-shot}     & 0.841 & 0.842 & 0.510 & 0.505 & 0.504 & 0.843 \\
& \textbf{Few-shot}      & \textbf{0.868} & \textbf{0.861} & 0.516 & 0.516 & 0.516 & \textbf{0.868} \\
\cmidrule{2-8}
& \textbf{Few-shot + CR} & 0.782 & 0.782 & \textbf{0.645} & \textbf{0.563} & \textbf{0.587} & 0.816 \\
\midrule

\multirow{3}{*}{\textbf{SE}} 

& \textbf{Zero-shot}     & 0.408 & 0.409 & 0.294 & 0.245 & 0.200 & 0.333 \\
& \textbf{Few-shot}      & \textbf{0.420} & \textbf{0.421} & 0.348 & 0.252 & 0.205 & \textbf{0.341} \\
\cmidrule{2-8}
& \textbf{Few-shot + CR} & 0.416 & 0.416 & \textbf{0.435} & \textbf{0.312} & \textbf{0.255} & 0.340 \\
\midrule

\multirow{3}{*}{\textbf{SI}} 

& \textbf{Zero-shot}     & 0.384 & 0.377 & \textbf{0.571} & 0.377 & 0.338 & 0.345 \\
& \textbf{Few-shot}      & \textbf{0.414} & \textbf{0.409} & 0.535 & \textbf{0.409} & \textbf{0.368} & \textbf{0.375} \\
\cmidrule{2-8}
& \textbf{Few-shot + CR} & 0.389 & 0.389 & 0.446 & 0.324 & 0.300 & 0.360 \\
\bottomrule
\end{tabular}
\end{threeparttable}
\end{table}

\begin{table}[ht]
\centering
\footnotesize
\begin{threeparttable}
\caption{Performance on Single Violated Rule Identification ( Gemini 2.5 Pro)}
\label{tab:medium_results_gemini25pro}
\setlength{\tabcolsep}{2pt}
\begin{tabular}{llcccccc}
\toprule
\textbf{Statement} & \textbf{Method} & \textbf{Acc} & \textbf{Bal. Acc} & \textbf{M-Prec} & \textbf{M-Rec} & \textbf{M-F1} & \textbf{W-F1} \\
\midrule

\multirow{3}{*}{\textbf{BS}} 

& \textbf{Zero-shot}     & 0.472 & 0.611 & 0.730 & 0.611 & 0.616 & \textbf{0.691} \\
& \textbf{Few-shot}      & 0.428 & 0.674 & 0.708 & 0.674 & 0.652 & 0.671 \\
\cmidrule{2-8}
& \textbf{Few-shot + CR} & \textbf{0.491} & \textbf{0.729} & \textbf{0.759} & \textbf{0.729} & \textbf{0.666} & 0.668 \\
\midrule

\multirow{3}{*}{\textbf{CF}} 

& \textbf{Zero-shot}     & 0.937 & 0.938 & \textbf{1.000} & 0.938 & \textbf{0.968} & \textbf{0.968} \\
& \textbf{Few-shot}      & \textbf{0.943} & \textbf{0.943} & 0.926 & \textbf{0.943} & 0.944 & 0.944 \\
\cmidrule{2-8}
& \textbf{Few-shot + CR} & 0.664 & 0.743 & 0.520 & 0.743 & 0.726 & 0.818 \\
\midrule

\multirow{3}{*}{\textbf{SE}} 

& \textbf{Zero-shot}     & 0.340 & 0.549 & 0.590 & 0.549 & 0.552 & 0.566 \\
& \textbf{Few-shot}      & 0.326 & 0.526 & 0.458 & 0.526 & 0.476 & 0.488 \\
\cmidrule{2-8}
& \textbf{Few-shot + CR} & \textbf{0.346} & \textbf{0.603} & \textbf{0.642} & \textbf{0.603} & \textbf{0.608} & \textbf{0.612} \\
\midrule

\multirow{3}{*}{\textbf{SI}} 

& \textbf{Zero-shot}     & 0.339 & 0.339 & 0.191 & 0.339 & 0.232 & 0.232 \\
& \textbf{Few-shot}      & \textbf{0.342} & 0.345 & 0.183 & 0.345 & 0.226 & 0.246 \\
\cmidrule{2-8}
& \textbf{Few-shot + CR} & 0.333 & \textbf{0.372} & \textbf{0.245} & \textbf{0.372} & \textbf{0.298} & \textbf{0.325} \\
\bottomrule
\end{tabular}
\end{threeparttable}
\end{table}

\begin{table}[ht]
\centering
\footnotesize
\begin{threeparttable}
\caption{Performance on Single Violated Rule Identification (Gemini 2.0)}
\label{tab:medium_results_gemini}
\setlength{\tabcolsep}{2pt}
\begin{tabular}{llcccccc}
\toprule
\textbf{Statement} & \textbf{Method} & \textbf{Acc} & \textbf{Bal. Acc} & \textbf{M-Prec} & \textbf{M-Rec} & \textbf{M-F1} & \textbf{W-F1} \\
\midrule

\multirow{3}{*}{\textbf{BS}} 

& \textbf{Zero-shot}     & 0.658 & 0.659 & \textbf{0.819} & 0.659 & 0.620 & 0.621 \\
& \textbf{Few-shot}      & 0.678 & 0.678 & 0.694 & 0.678 & 0.618 & 0.618 \\
\cmidrule{2-8}
& \textbf{Few-shot + CR} & \textbf{0.704} & \textbf{0.708} & 0.782 & \textbf{0.708} & \textbf{0.665} & \textbf{0.664} \\
\midrule

\multirow{3}{*}{\textbf{CF}} 

& \textbf{Zero-shot}     & 0.862 & 0.862 & 0.883 & 0.862 & 0.864 & 0.864 \\
& \textbf{Few-shot}      & 0.861 & 0.861 & 0.890 & 0.861 & 0.860 & 0.860 \\
\cmidrule{2-8}
& \textbf{Few-shot + CR} & \textbf{0.873} & \textbf{0.873} & \textbf{0.912} & \textbf{0.873} & \textbf{0.887} & \textbf{0.887} \\
\midrule

\multirow{3}{*}{\textbf{SE}} 

& \textbf{Zero-shot}     & 0.360 & 0.360 & \textbf{0.667} & 0.360 & 0.232 & 0.232 \\
& \textbf{Few-shot}      & 0.345 & 0.345 & 0.518 & 0.345 & 0.200 & 0.200 \\
\cmidrule{2-8}
& \textbf{Few-shot + CR} & \textbf{0.362} & \textbf{0.362} & 0.608 & \textbf{0.362} & \textbf{0.240} & \textbf{0.240} \\
\midrule

\multirow{3}{*}{\textbf{SI}} 

& \textbf{Zero-shot}     & 0.378 & 0.378 & 0.213 & 0.378 & 0.259 & 0.259 \\
& \textbf{Few-shot}      & 0.381 & 0.382 & 0.204 & 0.382 & 0.252 & 0.275 \\
\cmidrule{2-8}
& \textbf{Few-shot + CR} & \textbf{0.407} & \textbf{0.409} & \textbf{0.240} & \textbf{0.409} & \textbf{0.280} & \textbf{0.310} \\
\bottomrule
\end{tabular}
\end{threeparttable}
\end{table}

\begin{table}[ht]
\centering
\footnotesize
\begin{threeparttable}
\caption{Performance on Single Violated Rule Identification (LLaMA-3)}
\label{tab:medium_results_llama3_final}
\setlength{\tabcolsep}{2pt}
\begin{tabular}{llcccccc}
\toprule
\textbf{Statement} & \textbf{Method} & \textbf{Acc} & \textbf{Bal. Acc} & \textbf{M-Prec} & \textbf{M-Rec} & \textbf{M-F1} & \textbf{W-F1} \\
\midrule

\multirow{3}{*}{\textbf{BS}} 

& \textbf{Zero-shot}     & 0.540 & 0.702 & 0.781 & 0.702 & 0.707 & 0.748 \\
& \textbf{Few-shot}      & \textbf{0.617} & \textbf{0.763} & \textbf{0.804} & \textbf{0.763} & \textbf{0.768} & \textbf{0.789} \\
\cmidrule{2-8}
& \textbf{Few-shot + CR} & 0.475 & 0.644 & 0.733 & 0.644 & 0.658 & 0.671 \\
\midrule

\multirow{3}{*}{\textbf{CF}} 

& \textbf{Zero-shot}     & \textbf{0.809} & \textbf{0.894} & \textbf{0.923} & \textbf{0.894} & \textbf{0.902} & \textbf{0.904} \\
& \textbf{Few-shot}      & 0.687 & 0.814 & 0.842 & 0.814 & 0.819 & 0.821 \\
\cmidrule{2-8}
& \textbf{Few-shot + CR} & 0.699 & 0.823 & 0.861 & 0.823 & 0.830 & 0.833 \\
\midrule

\multirow{3}{*}{\textbf{SE}} 

& \textbf{Zero-shot}     & 0.331 & 0.497 & 0.541 & 0.497 & 0.503 & 0.518 \\
& \textbf{Few-shot}      & 0.332 & 0.499 & 0.547 & 0.499 & 0.505 & 0.520 \\
\cmidrule{2-8}
& \textbf{Few-shot + CR} & \textbf{0.333} & \textbf{0.500} & \textbf{0.552} & \textbf{0.500} & \textbf{0.506} & \textbf{0.522} \\
\midrule

\multirow{3}{*}{\textbf{SI}} 

& \textbf{Zero-shot}     & \textbf{0.396} & \textbf{0.567} & \textbf{0.601} & \textbf{0.567} & \textbf{0.571} & \textbf{0.584} \\
& \textbf{Few-shot}      & 0.391 & 0.562 & 0.596 & 0.562 & 0.566 & 0.579 \\
\cmidrule{2-8}
& \textbf{Few-shot + CR} & 0.377 & 0.547 & 0.583 & 0.547 & 0.552 & 0.565 \\
\bottomrule
\end{tabular}
\end{threeparttable}
\end{table}

\begin{table}[ht]
\centering
\footnotesize
\begin{threeparttable}
\caption{Performance Comparison of GPT-4o on Financial Statement Joint Multi-Step Diagnosis}
\label{tab:hard_results}
\setlength{\tabcolsep}{2pt}
\begin{tabular}{llcccccccc}
\toprule
\multirow{3}{*}{\textbf{Task}} & \multirow{3}{*}{\textbf{Method}} 
& \multicolumn{4}{c}{\textbf{Step 1 Violation}} 
& \multicolumn{4}{c}{\textbf{Violation Classification}} \\
& 
& \multicolumn{4}{c}{\textbf{Detection (Yes/No)}} 
& \multicolumn{4}{c}{\textbf{(Multi-label)}} \\
\cmidrule(lr){3-6} \cmidrule(lr){7-10}
& & \textbf{Acc} & \textbf{Prec} & \textbf{Rec} & \textbf{F1} 
& \textbf{EM} & \textbf{Prec$_\mu$} & \textbf{Rec$_\mu$} & \textbf{F1$_\mu$} \\
\midrule

\multirow{3}{*}{\textbf{BS}} 

& \textbf{Zero-shot}     & 77.0 & 79.4 & \textbf{96.0} & 86.91 & 15.0 & 49.3 & \textbf{84.7} & 62.3 \\
& \textbf{Few-shot}      & 76.0 & 80.1 & 93.5 & 86.3 & 18.0 & 52.1 & 78.3 & \textbf{62.6} \\
\cmidrule{2-10}
& \textbf{Few-shot + CR} & \textbf{78.0} & \textbf{82.3} & 92.7 & \textbf{87.2} & \textbf{24.0} & \textbf{59.3} & 66.1 & 62.5 \\
\midrule

\multirow{3}{*}{\textbf{CF}} 

& \textbf{Zero-shot}     & 88.0 & 90.2 & \textbf{96.0} & \textbf{93.0} & \textbf{43.0} & 89.4 & \textbf{72.1} & \textbf{79.8} \\
& \textbf{Few-shot}      & 89.0 & \textbf{98.2} & 88.3 & \textbf{92.9} & 40.0 & 91.2 & 64.5 & 75.5 \\
\cmidrule{2-10}
& \textbf{Few-shot + CR} & \textbf{91.0} & 97.1 & 91.9 & \textbf{94.4} & \textbf{43.0} & \textbf{95.6} & 65.8 & 77.9 \\
\midrule

\multirow{3}{*}{\textbf{SE}} 

& \textbf{Zero-shot}     & 79.0 & 81.4 & \textbf{97.5} & 88.7 & 20.0 & 60.6 & \textbf{93.2} & 73.4 \\
& \textbf{Few-shot}      & \textbf{81.0} & \textbf{83.2} & 96.0 & \textbf{89.1} & 29.0 & \textbf{66.5} & 85.2 & \textbf{74.7} \\
\cmidrule{2-10}
& \textbf{Few-shot + CR} & \textbf{81.0} & 81.7 & \textbf{98.1} & \textbf{89.1} & \textbf{30.0} & 65.6 & 86.4 & 74.5 \\
\midrule

\multirow{3}{*}{\textbf{SI}} 

& \textbf{Zero-shot}     & \textbf{85.0} & \textbf{90.0} & \textbf{91.1} & \textbf{90.5} & \textbf{33.0} & \textbf{69.2} & \textbf{65.1} & \textbf{67.1} \\
& \textbf{Few-shot}      & 81.0 & 87.4 & 88.1 & 87.7 & 26.0 & 66.5 & 62.1 & 64.2 \\
\cmidrule{2-10}
& \textbf{Few-shot + CR} & 83.0 & 89.6 & 90.5 & 90.1 & \textbf{33.0} & 66.9 & 67.0 & 66.9 \\
\bottomrule
\end{tabular}
\end{threeparttable}
\end{table}

\begin{table}[ht]
\centering
\footnotesize
\begin{threeparttable}
\caption{Performance Comparison of Gemini 2.5 Pro on Financial Statement Joint Multi-Step Diagnosis}
\label{tab:gemini_pro_results}
\setlength{\tabcolsep}{2pt}
\begin{tabular}{llcccccccc}
\toprule
\multirow{3}{*}{\textbf{Task}} & \multirow{3}{*}{\textbf{Method}} 
& \multicolumn{4}{c}{\textbf{Step 1 Violation}} 
& \multicolumn{4}{c}{\textbf{Violation Classification}} \\
& 
& \multicolumn{4}{c}{\textbf{Detection (Yes/No)}} 
& \multicolumn{4}{c}{\textbf{(Multi-label)}} \\
\cmidrule(lr){3-6} \cmidrule(lr){7-10}
& & \textbf{Acc} & \textbf{Prec} & \textbf{Rec} & \textbf{F1} 
& \textbf{EM} & \textbf{Prec$_\mu$} & \textbf{Rec$_\mu$} & \textbf{F1$_\mu$} \\
\midrule

\multirow{3}{*}{\textbf{BS}} 

& \textbf{Zero-shot}     & 81.0 & 80.8 & \textbf{100.0} & 89.4 & 26.0 & 63.6 & \textbf{85.0} & 72.7 \\
& \textbf{Few-shot}      & \textbf{83.0} & \textbf{82.5} & \textbf{100.0} & \textbf{90.4} & \textbf{35.0} & \textbf{66.5} & 84.4 & \textbf{74.4} \\
\cmidrule{2-10}
& \textbf{Few-shot + CR} & 81.0 & 80.8 & \textbf{100.0} & 89.4 & 28.0 & 59.3 & 83.8 & 69.4 \\
\midrule

\multirow{3}{*}{\textbf{CF}} 

& \textbf{Zero-shot}     & 84.0 & 83.8 & \textbf{100.0} & 91.2 & \textbf{57.0} & 81.5 & \textbf{90.4} & \textbf{85.7} \\
& \textbf{Few-shot}      & 87.0 & 89.8 & 95.2 & 92.4 & 49.0 & 89.9 & 74.7 & 81.6 \\
\cmidrule{2-10}
& \textbf{Few-shot + CR} & \textbf{91.0} & \textbf{96.3} & 92.8 & \textbf{94.5} & 39.0 & \textbf{96.3} & 62.0 & 75.5 \\
\midrule

\multirow{3}{*}{\textbf{SE}} 

& \textbf{Zero-shot}     & 86.0 & 86.8 & \textbf{97.5} & 91.9 & 55.0 & 82.2 & 82.7 & 82.5 \\
& \textbf{Few-shot}      & \textbf{90.0} & \textbf{90.8} & \textbf{97.5} & \textbf{94.0} & \textbf{62.0} & \textbf{85.2} & \textbf{85.2} & \textbf{85.2} \\
\cmidrule{2-10}
& \textbf{Few-shot + CR} & 88.0 & 88.8 & \textbf{97.5} & 92.9 & 56.0 & 82.6 & 82.1 & 82.4 \\
\midrule

\multirow{3}{*}{\textbf{SI}} 

& \textbf{Zero-shot}     & \textbf{83.0} & \textbf{85.1} & 94.9 & 89.7 & \textbf{37.0} & \textbf{65.9} & 75.6 & \textbf{70.4} \\
& \textbf{Few-shot}      & 81.0 & 83.9 & 93.6 & 88.5 & 36.0 & 64.2 & 76.9 & 70.0 \\
\cmidrule{2-10}
& \textbf{Few-shot + CR} & \textbf{83.0} & 84.3 & \textbf{96.2} & \textbf{89.8} & 34.0 & 61.7 & \textbf{81.4} & 70.2 \\
\bottomrule
\end{tabular}
\begin{tablenotes}
\small
\item \textit{Note:} All metrics reported as percentages. \textbf{CR} = Counterfactual Reasoning. \textbf{EM} = Exact Match (strict match of all violation codes). Prec$_\mu$/Rec$_\mu$/F1$_\mu$ denote micro-averaged metrics across all violation types.
\end{tablenotes}
\end{threeparttable}
\end{table}

\begin{table}[ht]
\centering
\footnotesize
\begin{threeparttable}
\caption{Performance Comparison of Gemini 2.0 flash on Financial Statement Joint Multi-Step Diagnosis}
\label{tab:gemini_results}
\setlength{\tabcolsep}{2pt}
\begin{tabular}{llcccccccc}
\toprule
\multirow{3}{*}{\textbf{Task}} & \multirow{3}{*}{\textbf{Method}} 
& \multicolumn{4}{c}{\textbf{Step 1 Violation}} 
& \multicolumn{4}{c}{\textbf{Violation Classification}} \\
& 
& \multicolumn{4}{c}{\textbf{Detection (Yes/No)}} 
& \multicolumn{4}{c}{\textbf{(Multi-label)}} \\
\cmidrule(lr){3-6} \cmidrule(lr){7-10}
& & \textbf{Acc} & \textbf{Prec} & \textbf{Rec} & \textbf{F1} 
& \textbf{EM} & \textbf{Prec$_\mu$} & \textbf{Rec$_\mu$} & \textbf{F1$_\mu$} \\
\midrule

\multirow{3}{*}{\textbf{BS}} 

& \textbf{Zero-shot}     & 78.0 & \textbf{82.2} & 92.5 & 87.1 & 21.0 & 59.8 & 69.8 & 63.9 \\
& \textbf{Few-shot}      & \textbf{81.0} & 82.1 & \textbf{97.5} & \textbf{89.1} & \textbf{29.0} & \textbf{65.1} & \textbf{81.9} & \textbf{72.5} \\
\cmidrule{2-10}
& \textbf{Few-shot + CR} & 77.0 & 80.6 & 93.7 & 86.7 & 20.0 & 55.9 & 70.6 & 62.4 \\
\midrule

\multirow{3}{*}{\textbf{CF}} 

& \textbf{Zero-shot}     & 90.0 & 93.9 & \textbf{93.9} & 93.9 & 40.0 & 93.8 & 63.8 & 75.9 \\
& \textbf{Few-shot}      & \textbf{91.0} & \textbf{95.1} & \textbf{93.9} & \textbf{94.5} & \textbf{47.0} & 94.1 & \textbf{68.0} & \textbf{79.0} \\
\cmidrule{2-10}
& \textbf{Few-shot + CR} & 85.0 & 94.7 & 86.7 & 90.5 & 41.0 & \textbf{94.2} & 59.6 & 73.0 \\
\midrule

\multirow{3}{*}{\textbf{SE}} 

& \textbf{Zero-shot}     & 68.0 & 79.5 & 81.5 & 80.5 & 26.0 & 62.4 & 79.5 & 68.9 \\
& \textbf{Few-shot}      & \textbf{81.0} & \textbf{82.2} & \textbf{97.5} & 82.2 & \textbf{34.0} & 65.9 & \textbf{93.2} & \textbf{77.2} \\
\cmidrule{2-10}
& \textbf{Few-shot + CR} & 76.0 & 81.3 & 91.3 & \textbf{86.0} & 30.0 & \textbf{69.2} & 72.2 & 70.7 \\
\midrule

\multirow{3}{*}{\textbf{SI}} 

& \textbf{Zero-shot}     & 82.0 & \textbf{89.4} & 87.1 & 88.3 & \textbf{25.0} & \textbf{69.0} & 55.7 & 61.7 \\
& \textbf{Few-shot}      & \textbf{83.0} & 85.8 & \textbf{93.5} & \textbf{89.5} & 23.0 & 62.1 & 67.3 & \textbf{64.6} \\
\cmidrule{2-10}
& \textbf{Few-shot + CR} & 81.0 & 83.9 & \textbf{93.5} & 88.5 & 22.0 & 55.7 & \textbf{69.2} & 61.7 \\
\bottomrule
\end{tabular}
\begin{tablenotes}
\small
\item \textit{Note:} All metrics reported as percentages. \textbf{CR} = Counterfactual Reasoning. \textbf{EM} = Exact Match (strict match of all violation codes). Prec$_\mu$/Rec$_\mu$/F1$_\mu$ denote micro-averaged metrics across all violation types. 
\end{tablenotes}
\end{threeparttable}
\end{table}

\begin{table}[ht]
\centering
\footnotesize
\begin{threeparttable}
\caption{Performance Comparison of LLaMA-3 on Financial Statement Joint Multi-Step Diagnosis}
\label{tab:llama3_results}
\setlength{\tabcolsep}{2pt}
\begin{tabular}{llcccccccc}
\toprule
\multirow{3}{*}{\textbf{Task}} & \multirow{3}{*}{\textbf{Method}} 
& \multicolumn{4}{c}{\textbf{Step 1 Violation}} 
& \multicolumn{4}{c}{\textbf{Violation Classification}} \\
& 
& \multicolumn{4}{c}{\textbf{Detection (Yes/No)}} 
& \multicolumn{4}{c}{\textbf{(Multi-label)}} \\
\cmidrule(lr){3-6} \cmidrule(lr){7-10}
& & \textbf{Acc} & \textbf{Prec} & \textbf{Rec} & \textbf{F1} 
& \textbf{EM} & \textbf{Prec$_\mu$} & \textbf{Rec$_\mu$} & \textbf{F1$_\mu$} \\
\midrule

\multirow{3}{*}{\textbf{BS}} 

& \textbf{Zero-shot}     & 80.0 & 80.0 & \textbf{100.0} & 88.9 & 8.0 & 45.6 & 84.4 & 59.2 \\
& \textbf{Few-shot}      & \textbf{84.0} & \textbf{84.8} & 97.5 & \textbf{90.7} & \textbf{16.0} & 45.7 & \textbf{91.9} & \textbf{61.0} \\
\cmidrule{2-10}
& \textbf{Few-shot + CR} & 73.0 & 82.7 & 83.8 & 83.2 & 14.0 & \textbf{47.9} & 48.8 & 48.3 \\
\midrule

\multirow{3}{*}{\textbf{CF}} 

& \textbf{Zero-shot}     & \textbf{83.0} & 83.0 & \textbf{100.0} & \textbf{90.7} & \textbf{53.0} & 79.0 & \textbf{92.8} & \textbf{85.3} \\
& \textbf{Few-shot}      & 81.0 & 83.5 & 95.0 & 88.9 & 11.0 & 45.7 & 89.4 & 60.5 \\
\cmidrule{2-10}
& \textbf{Few-shot + CR} & 82.0 & \textbf{100.0} & 78.3 & 87.8 & 28.0 & \textbf{76.3} & 63.9 & 69.5 \\
\midrule

\multirow{3}{*}{\textbf{SE}} 

& \textbf{Zero-shot}     & 34.0 & 74.2 & 28.4 & 41.1 & 11.0 & 58.5 & 19.1 & 28.8 \\
& \textbf{Few-shot}      & \textbf{55.0} & \textbf{92.9} & \textbf{48.1} & \textbf{63.4} & \textbf{20.0} & \textbf{62.4} & \textbf{35.8} & \textbf{45.5} \\
\cmidrule{2-10}
& \textbf{Few-shot + CR} & 31.0 & 87.5 & 17.3 & 28.9 & 17.0 & 56.5 & 8.0 & 14.1 \\
\midrule

\multirow{3}{*}{\textbf{SI}} 

& \textbf{Zero-shot}     & \textbf{81.0} & \textbf{88.3} & \textbf{87.2} & \textbf{87.7} & \textbf{16.0} & 48.2 & \textbf{69.2} & \textbf{56.8} \\
& \textbf{Few-shot}      & 79.0 & 87.1 & 86.2 & \textbf{87.7} & 15.0 & 49.0 & 66.0 & 56.3 \\
\cmidrule{2-10}
& \textbf{Few-shot + CR} & 79.0 & 88.0 & 84.6 & 86.3 & 14.0 & \textbf{58.2} & 52.6 & 55.2 \\
\bottomrule
\end{tabular}
\begin{tablenotes}
\small
\item \textit{Note:} All metrics reported as percentages. \textbf{CR} = Counterfactual Reasoning. \textbf{EM} = Exact Match (strict match of all violation codes). Prec$_\mu$/Rec$_\mu$/F1$_\mu$ denote micro-averaged metrics across all violation types.
\end{tablenotes}
\end{threeparttable}
\end{table}



\begin{table}[ht]
\centering
\footnotesize
\begin{threeparttable}
\caption{Error Type Dist. for Joint Rule Diagnosis under Few-shot + Counterfactual Prompting (GPT-4o).}
\label{tab:hard_error_type_all}
\setlength{\tabcolsep}{2pt}
\begin{tabular}{lccccc}
\toprule
\multirow{2}{*}{\textbf{Statement}} & \multirow{2}{*}{\textbf{Correct}} & \multicolumn{4}{c}{\textbf{Error Breakdown}} \\
\cmidrule(lr){3-6}
 & & \textbf{Step-1 Error} & \textbf{Partial Det.} & \textbf{Partial Hall.} & \textbf{Inc. Loc.} \\
\midrule

\textbf{BS} & 18\% & 22\% & 10\% & 41\% & 9\% \\
\textbf{CF} & 41\% & 9\% & 35\% & 4\% & 11\% \\
\textbf{SE} & 29\% & 19\% & 10\% & 34\% & 8\% \\
\textbf{SI} & 20\% & 18\% & 22\% & 31\% & 9\% \\

\bottomrule
\end{tabular}
\begin{tablenotes}
\small
\item \textit{Note:} Results are computed at the instance level for the joint rule diagnosis task under few-shot + causal-counterfactual prompting.
\end{tablenotes}
\end{threeparttable}
\end{table}

\newcommand{\autowrap}[1]{\parbox{2.2cm}{\raggedright\strut#1}}
\newcolumntype{L}{>{\raggedright\arraybackslash}X}
\newcolumntype{C}{>{\centering\arraybackslash}p{1.5cm}}
\setlength{\tabcolsep}{4pt}
\newcommand{\multilinecell}[2][c]{\multirow{#2}{2.2cm}{\makecell[#1]{\raggedright\strut#1}}}

\onecolumn
\begin{longtable}{@{}>{\raggedright}p{1.4cm}p{2.0cm}p{3.3cm}Cp{6.5cm}@{}}
\caption{Financial Statement Rules Example} \\
\toprule
\textbf{Category} & \textbf{Rule ID} & \textbf{Rule Description} & \textbf{FASB Reference} & \textbf{FASB Guideline Text} \\
\midrule
\endfirsthead

\midrule
\textbf{Category} & \textbf{Rule ID} & \textbf{Rule Description} & \textbf{FASB Reference} & \textbf{FASB Guideline Text} \\
\midrule
\endhead

\bottomrule
\endfoot

\multirow{5}{*}{\autowrap{I. Balance Sheet Rules}}
 & BS-01 & Total Assets = Total Liabilities + Total Ownership Interests (including parent company common and preferred equity, non-controlling interests, and any equity-classified mezzanine items).
 & FASB Concepts Statement No. 6 (CON 6)
 & Equity or net assets is the residual interest in the assets of an entity that remains after deducting its liabilities. In a business enterprise, the equity is the ownership interest. \\
\addlinespace[0.2cm]

 & BS-02 & Assets and liabilities must be classified into current and non-current categories.
 & 210-10-45-5/210-10-05-4
 & A total of current liabilities shall be presented in classified balance sheets./ The balance sheets of most entities show separate classifications of current assets and current liabilities (commonly referred to as classified balance sheets) permitting ready determination of working capital. \\
\addlinespace[0.2cm]

 & BS-03 & The statement of cash flows must use descriptive terms like "cash" or "cash and cash equivalents".
 & 230-10-45-4 
 & The statement shall use descriptive terms such as cash or cash and cash equivalents rather than ambiguous terms such as funds. \\
\addlinespace[0.2cm]

 & BS-04 & Any appropriation of retained earnings must be presented separately within the shareholders' equity section. 
 & 505-10-45-3 
 & Appropriation of retained earnings is permitted, provided that it is shown within the shareholders' equity section of the balance sheet and is clearly identified as an appropriation of retained earnings. \\
\addlinespace[0.2cm]

 & BS-05 & Treasury stock should be reported as a deduction from total shareholders' equity. 
 & 505-10-05-1 
 & The Equity Topic includes the following Subtopics: a. Overall b. Stock Dividends and Stock Splits c. Treasury Stock d. Subparagraph superseded by Accounting Standards Update No. 2018-07. e. Spinoffs and Reverse Spinoffs. \\
\addlinespace[0.4cm]

\multirow{5}{*}{\autowrap{II. Income Statement Rules}} 
 & SI-01 & The results of operating activities must be reported separately from non-operating activities. 
 & 255-10-50-3
 & An entity shall disclose all of the following information for each of the five most recent years: a. Net sales and other operating revenues b. Income from continuing operations on a current cost basis c. ......
\\
\addlinespace[0.2cm]

 & SI-02 & Depreciation expense for the period must be included in operating expenses. 
 & 360-10-35-4 
 & The cost of a productive facility is one of the costs of the services it renders during its useful economic life. Generally accepted accounting principles (GAAP) require that this cost be spread over the expected useful life of the facility in ...... \\
\addlinespace[0.2cm]

 & SI-03 & Earnings per share (EPS) for both basic and diluted shares must be presented on the face of the income statement.
 & 260-10-45-2 
 & Entities with simple capital structures, that is, those with only common stock outstanding, shall present basic per-share amounts for income from continuing operations and for net income on the face of the income statement. ...... \\
\addlinespace[0.2cm]

 & SI-04 & On the income statement, a material gain from a transaction outside of primary business operations must be presented as a distinct line item, with the nature of that transaction clearly disclosed.
 & 220-10-S99-2
 & Material amounts included under miscellaneous other income shall be separately stated in the statement of comprehensive income or in a note thereto, indicating clearly the nature of the transactions out of which the items arose. \\
\addlinespace[0.2cm]

 & SI-05 & An event or transaction that is material and either unusual in nature or infrequent in occurrence must be presented as a separate component of income from continuing operations, and must not be reported on the face of the income statement net of income taxes.
 & 220-20-45-1
 & A material event or transaction that an entity considers to be of an unusual nature or of a type that indicates infrequency of occurrence or both shall be reported as a separate component of income from continuing operations........ \\
\addlinespace[0.4cm]

\multirow{5}{*}{\autowrap{III. Cash Flow Statement Rules}} 
 & CF-01 & Net Cash Flow from Operating Activities + Investing Activities + Financing Activities = Net Change in Cash and Cash Equivalents (note: consider the effect of exchange rate changes).
 & 230-10-45-24
 & A statement of cash flows for a period shall report net cash provided or used by operating, investing, and financing activities and the net effect of those flows on the total of cash, cash equivalents, .....
\\
\addlinespace[0.2cm]

 & CF-02 & Financial statements shall not report an amount of cash flow per share.
 & 230-10-45-3 
 & Financial statements shall not report an amount of cash flow per share. Neither cash flow nor any component of it is an alternative to net income as an indicator of an entity's performance, as reporting per-share amounts might imply...... \\
\addlinespace[0.2cm]

 & CF-03 & Clearly show changes in cash, cash equivalents, and restricted cash in the cash-flow statement, use precise terms (not “funds”).
 & 230-10-45-4 
 & A statement of cash flows shall explain the change during the period in the total of cash, cash equivalents, and amounts generally described as restricted cash or restricted cash equivalents. The statement shall use...... \\
\addlinespace[0.4cm]

\multirow{5}{*}{\autowrap{IV. Statement of Equity Rules}} 

 & SE-01 & The Statement of Equity must follow certain rules, such as the following equation regarding the composition of Total Equity: Total Equity (year) = Common Stock and Additional Paid-in Capital + Retained Earnings + Accumulated Other Comprehensive Income + Non-controlling Interest + Treasury Stock.
 & 505-10-50-2
 & If both financial position and results of operations are presented, disclosure of changes in the separate accounts comprising shareholders' equity (in addition to retained earnings) and of the changes in the number of shares of equity securities during at least the most recent annual fiscal period and any subsequent interim period presented is required to make the financial statements sufficiently informative. Disclosure of such changes may take the form of separate statements or may be made in the basic financial statements or notes thereto.
\\
\addlinespace[0.2cm]

 & SE-02 & Net income (or loss) for the period must be shown as an addition (or deduction) to retained earnings.
 & 215-10-50-1
 & If both financial position and results of operations are presented, disclosure of changes in the separate accounts comprising shareholders' equity (in addition to retained earnings) and of the changes in the number of shares of equity securities during at least the most recent annual fiscal period and any subsequent interim period presented is required to make the financial statements sufficiently informative. Disclosure of such changes may take the form of separate statements or may be made in the basic financial statements or notes thereto.
\\
\addlinespace[0.2cm]

 & SE-03 & Other comprehensive income (OCI) items must be presented as changes to accumulated other comprehensive income (AOCI).
 & 220-10-45-14
 & The total of other comprehensive income for a period shall be transferred to a component of equity that is presented separately from retained earnings and additional paid-in capital in a statement of financial position at the end of an accounting period. A descriptive title such as accumulated other comprehensive income shall be used for that component of equity.
\\

\label{table:acc_rules} 
\end{longtable}

\begin{multicols}{2}
\section{Causal-Counterfactual Prompting Details}
\label{app:cc_prompts}

This appendix provides the exact prompt templates and representative screenshots used for causal-counterfactual prompting in FinRule-Bench. These materials are included for reproducibility and transparency. The main paper describes the prompting protocol at a high level in Section~3.

\subsection{Prompt Template Overview}

Causal-counterfactual prompting is implemented using in-context exemplars that explicitly specify (i) the causal source of a rule violation and (ii) a minimal counterfactual modification that would restore compliance. The templates below show the exact prompt structure used across tasks. Each exemplar contains a financial table, a target decision, a causal explanation identifying the violated principle, and a minimal counterfactual edit that would change the validator outcome. Exemplars are fixed across models and tasks to ensure comparability.

\subsection{Representative Prompt}
\end{multicols}

\begin{tcolorbox}[colback=gray!5!white, colframe=black!80!black,
title=Representative causal-counterfactual prompt for the \textbf{Rule Verification} task (single violated rule identification).,
boxrule=0.4mm, width=\linewidth, arc=1.6mm, auto outer arc=true]
\small
\textbf{System Instruction:}
You are a strict financial auditor evaluating the following accounting rule.

\textbf{Rule Definition (BS-01):}
Total Assets MUST equal (Total Liabilities + Total Equity + Mezzanine Equity).

\textbf{Examples:}
\begin{itemize}[leftmargin=*, topsep=0pt, itemsep=2pt]
    \item \textbf{COMPLIANT $\rightarrow$ T} \\
    $\text{Assets}: 42,448 = \text{Liabilities}: 35,451 + \text{Equity}: 6,975 + \text{Mezzanine}: 22$ \\
    \textit{CAUSE:} The accounting identity holds exactly. \\
    \textit{COUNTERFACTUAL:} If Assets were changed to 42,449, the identity would be violated.

    \item \textbf{NON-COMPLIANT $\rightarrow$ F} \\
    $\text{Assets}: 42,500 \neq \text{Liabilities}: 35,451 + \text{Equity}: 6,975 + \text{Mezzanine}: 22$ \\
    \textit{CAUSE:} Total Assets do not equal the sum of Liabilities, Equity, and Mezzanine. \\
    \textit{COUNTERFACTUAL:} If Assets were changed to 42,448, the identity would hold.
\end{itemize}

\textbf{Evaluation Task:}
EVALUATE THE FOLLOWING BALANCE SHEET:

BALANCE SHEET DATA:
\texttt{<INSERT MARKDOWN TABLE HERE>}

Respond with 'T' if compliant, 'F' if non-compliant. \\
Answer:
\end{tcolorbox}

\begin{tcolorbox}[colback=gray!5!white, colframe=black!80!black,
title=Representative causal-counterfactual prompt for the \textbf{Rule Identification} task (single violated rule identification)., boxrule=0.4mm, width=\linewidth, arc=1.6mm, auto outer arc=true]
\small
\textbf{System Instruction:}
You are a strict financial auditor.

\textbf{Task:}
Determine whether the balance sheet violates ANY accounting rules. If violations exist, report ALL violated rules.

\textbf{Rule Priority:}
Check BS01 first, then BS02 $\rightarrow$ BS03 $\rightarrow$ BS04 $\rightarrow$ BS05.

\textbf{Rules:}
\begin{itemize}[leftmargin=*, topsep=0pt, itemsep=2pt]
    \item \textbf{BS01:} Total Assets MUST equal (Total Liabilities + Total Equity).
    \item \textbf{BS02:} ALL asset items MUST be classified as "Current\_Assets" or "Non\_Current\_Assets".
    \item \textbf{BS03:} Cash MUST use exact term "Cash and cash equivalents".
    \item \textbf{BS04:} Retained earnings MUST be labeled "Retained earnings".
    \item \textbf{BS05:} Treasury stock (if any) MUST be in Equity as a NEGATIVE value.
\end{itemize}

\textbf{Reference Examples:}
\begin{itemize}[leftmargin=*, topsep=0pt, itemsep=2pt]
    \item \textbf{NON-COMPLIANT $\rightarrow$ YES: [BS03, BS04]} \\
    \textit{Data Context:}
    Assets | Current\_Assets | Funds and funds equivalents | 5000 \\
    Equity | Stockholders\_Equity | Accumulated profits | 10000 \\
    \textit{CAUSE:}
    - BS03 violated due to non-GAAP cash terminology ("Funds...").
    - BS04 violated because it says "Accumulated profits" instead of "Retained earnings" (non-standard label). \\
    \textit{COUNTERFACTUAL:}
    - Renaming "Funds" to "Cash and cash equivalents" resolves BS03.
    - Renaming "Accumulated profits" to "Retained earnings" resolves BS04.

    \item \textbf{COMPLIANT $\rightarrow$ NO} \\
    \textit{Condition:}
    BS01: $18000 = 8000 + 10000$ (Identity holds). \\
    BS02: Current\_Assets/Non\_Current\_Assets present. \\
    BS05: No treasury stock.
\end{itemize}

\textbf{Output Format:}
YES: [rule1, rule2, ...] \\
NO
\end{tcolorbox}

\begin{tcolorbox}[colback=gray!5!white, colframe=black!80!black,
title=Representative causal-counterfactual prompt for the \textbf{multi-rule diagnosis} task.,
boxrule=0.4mm, width=\linewidth, arc=1.6mm, auto outer arc=true]
\small
\textbf{System Instruction:}
You are a strict financial auditor.

\textbf{TASK:}
Determine whether the balance sheet violates ANY accounting rules. If violations exist, report ALL violated rules.

\textbf{RULE PRIORITY:}
Check BS01 first, then BS02 $\rightarrow$ BS03 $\rightarrow$ BS04 $\rightarrow$ BS05.

\textbf{RULES:}
\begin{itemize}[leftmargin=*, topsep=0pt, itemsep=2pt]
    \item \textbf{BS01:} Total Assets MUST equal (Total Liabilities + Total Equity).
    \item \textbf{BS02:} ALL asset items MUST be classified as "Current\_Assets" or "Non\_Current\_Assets".
    \item \textbf{BS03:} Cash MUST use exact term "Cash and cash equivalents".
    \item \textbf{BS04:} Retained earnings MUST be labeled "Retained earnings".
    \item \textbf{BS05:} Treasury stock (if any) MUST be in Equity as a NEGATIVE value.
\end{itemize}

\textbf{REFERENCE EXAMPLES:}
\begin{itemize}[leftmargin=*, topsep=0pt, itemsep=2pt]
    \item \textbf{Scenario 1: Violations Found} \\
    \textit{Input Data Snippet:} \\
    Assets | Current\_Assets | Funds and funds equivalents | 5000 \\
    Equity | Stockholders\_Equity | Accumulated profits | 10000 \\
    \textit{OUTPUT:} YES: [BS03, BS04] \\
    \textit{CAUSE:}
    - BS03 violated due to non-GAAP cash terminology.
    - Says "Accumulated profits" instead of "Retained earnings" (non-standard label). \\
    \textit{COUNTERFACTUAL:}
    - Renaming "Funds" to "Cash and cash equivalents" resolves BS03.
    - Renaming "Accumulated profits" to "Retained earnings" resolves BS04.

    \item \textbf{Scenario 2: Compliant} \\
    \textit{Condition:}
    BS01: $18000 = 8000 + 10000$ (Identity holds). \\
    BS02: Current\_Assets/Non\_Current\_Assets present. \\
    BS05: No treasury stock. \\
    \textit{OUTPUT:} NO
\end{itemize}

\textbf{OUTPUT FORMAT:}
YES: [rule1, rule2, ...] \\
NO
\end{tcolorbox}

\end{document}